\documentclass[journal]{IEEEtran}
\usepackage{amsmath,amsfonts}
\usepackage{algorithmic}
\usepackage{algorithm}
\usepackage{array}
\usepackage[caption=false,font=normalsize,labelfont=small,textfont=small]{subfig}
\usepackage{textcomp}
\usepackage{stfloats}
\usepackage{url}
\usepackage{verbatim}
\usepackage{graphicx}
\usepackage{cite}
\usepackage{multirow}
\usepackage{pifont}
\usepackage{threeparttable}
\usepackage{color}
\newcommand{\edita}[1]{\textcolor{black}{#1}}
\begin{document}

\title{DAM-Net: Domain Adaptation Network with Micro-Labeled Fine-Tuning for Change Detection}

\author{Hongjia Chen,
Xin~Xu,
Fangling Pu,~\IEEEmembership{Member,~IEEE}
\thanks{This work has been submitted to the IEEE for possible publication. Copyright may be transferred without notice, after which this version may no longer be accessible.}
\thanks{This work was supported in part by National Natural Science Foundation of China under Grant 62271356 and in part by National Natural Science Foundation of China under Grant 62071336. (\textit{Corresponding author: Xin Xu)}}
\thanks{The authors are with the Collaborative Sensing Laboratory, Electronic Information School, Wuhan University, Wuhan 430079, China (e-mail: chj1997@whu.edu.cn; xinxu@whu.edu.cn; flpu@whu.edu.cn).}}



\maketitle

\begin{abstract}
Change detection (CD) in remote sensing imagery plays a crucial role in various applications such as urban planning, damage assessment, and resource management. While deep learning approaches have significantly advanced CD performance, current methods suffer from poor domain adaptability, requiring extensive labeled data for retraining when applied to new scenarios. This limitation severely restricts their practical applications across different datasets. In this work, we propose DAM-Net: a Domain Adaptation Network with Micro-Labeled Fine-Tuning for CD. Our network introduces adversarial domain adaptation to CD for, utilizing a specially designed segmentation-discriminator and alternating training strategy to enable effective transfer between domains. Additionally, we propose a novel Micro-Labeled Fine-Tuning approach that strategically selects and labels a minimal amount of samples (less than 1\%) to enhance domain adaptation. The network incorporates a Multi-Temporal Transformer for feature fusion and optimized backbone structure based on previous research. Experiments conducted on the LEVIR-CD and WHU-CD datasets demonstrate that DAM-Net significantly outperforms existing domain adaptation methods, achieving comparable performance to semi-supervised approaches that require 10\% labeled data while using only 0.3\% labeled samples. Our approach significantly advances cross-dataset CD applications and provides a new paradigm for efficient domain adaptation in remote sensing.
The source code of DAM-Net will be made publicly available upon publication.
\end{abstract}


\section{Introduction}
\IEEEPARstart{C}{hange} detection (CD) aims to identify change areas in a given area by analyzing multiple remote sensing observations of the area at different times \cite{singh1989review,bruzzone2012novel}, and has gradually become a research hotspot in remote sensing \cite{yang2021da2net,li2020amn}.
CD models take multi-temporal images of the same region as input, and the change areas can be extracted pixel by pixel by comparison.
With the increasing availability of high-resolution remote sensing imagery, CD has been widely applied in various fields, including damage assessment \cite{lei2019landslide,peng2020optical}, urban planning \cite{ji2018fully,xian2010updating}, resource management \cite{song2014remote}, etc. \cite{wu2023fully,chen2019change,li2022transunetcd}

In recent years, the advancement of deep neural networks, including Convolutional Neural Networks (CNNs) \cite{lecun1998gradient}, Transformers \cite{vaswani2017attention}, Mamba \cite{gu2023mamba}, etc., has driven significant progress in CD \cite{gu2022parameterization,elman1990finding,hochreiter1997long}.
Deep neural networks have been widely used in remote sensing domain tasks including CD, and have driven the continuous progress of the field \cite{lecun2015deep}.
CD has gradually evolved from traditional multi-stage pixel-based and object-based methods \cite{hussain2013change,lambin1994change,marchesi2009ica} to end-to-end deep learning approaches \cite{shi2020change}.
Meanwhile, the increasing ease of access to very high resolution remote sensing imagery has greatly facilitated the widespread use of data-driven deep neural networks \cite{hypersigma}.

Although current deep learning CD networks have achieved quite impressive results, they face a critical challenge: poor domain adaptability.
These networks, despite achieving impressive results, heavily rely on large-scale labeled datasets \cite{zhu2017deep} and can only be used in specifically trained scenarios.
They often perform poorly when applied to new scenarios with different sensor attributes, environmental conditions, or geographical locations.
Therefore, the networks requires to be retrained using the labels of the new scenario, a process that is typically time-consuming and labor-intensive.
With the development of Remote Sensing Foundation Model (RSFM), the aforementioned issues have been partly alleviated.
RSFMs can extract robust features from remote sensing images, but they still require extensive fine-tuning for specific downstream tasks such as CD \cite{guo2024skysense}, and the cost of training is still considerable, which limits the wide application of large models.
To tackle this challenge, unsupervised domain adaptation (DA) \cite{wang2018deep,sun2016deep} has emerged as a promising solution, aiming to transfer models trained in a source domain to an unlabeled target domain. 
(In remote sensing, these applications are sometimes referred to as semi-supervised DA \cite{tuia2016domain}.)
Specifically, labeled data in the source domain and unlabeled data in the target domain are used to transfer models that have been trained in the source domain to the target domain.
Existing DA approaches typically employ generative adversarial networks (GANs) \cite{goodfellow2020generative}, adversarial domain adaptation (ADA) \cite{tsai2018learning}, or feature alignment techniques \cite{roy2022uncertainty}.
However, the effectiveness of current methods is unsatisfactory, making them difficult to implement in a general way.
While many existing works strive for completely label-free adaptation, we observe that having access to a tiny portion of labeled samples from the target domain could be valuable. Such a pragmatic approach, which we call micro-labeling, may offer a more practical trade-off between annotation effort and model performance.

In order to tackle the previously mentioned concern, we focused on domain adaptation and micro-labeled fine-tuning, enabling networks trained on a single dataset to be effectively applied to other datasets.
First, we further optimize the backbone structure of the network based on our previous researches RDP-Net \cite{chen2022rdpnet} and SRC-Net \cite{chen2024srcnet}, introducing a Multi-Temporal Transformer (MT-Transformer) for multi-temporal feature fusion.
Second, we introduce adversarial domain adaptation into the CD task with a redesigned segmentation-discriminator structure and alternating training strategy specifically tailored for CD characteristics.
The discriminator learns to distinguish features from source and target domains, while adversarial learning reduces the domain discrepancy, thereby achieving unsupervised domain adaptation.
Third, we introduce micro-labeling to CD tasks, where we utilize the discriminator to identify and select challenging samples from the target domain. By manually annotating only a minimal set of data (less than 1\%, typically 10-20 images) and fine-tuning the network with these samples, we can further enhance the model's adaptability to the target domain.
We believe that the domain adaptive approach proposed in our work will greatly broaden the application of CD across datasets.

The DAM-Net has achieved a better performance in the field of remote sensing CD. The major contributions of DAM-Net can be summarized as follows.

\begin{enumerate}
\item{We introduce adversarial domain adaptation to CD
with specifically designed components: a CD-oriented segmentation-discriminator and an alternating training strategy. This enables effective knowledge transfer from source to target domains without requiring target domain labels.}
\item{We propose Micro-Labeled Fine-Tuning,
a practical approach where we leverage the domain discriminator to identify and annotate only the most informative samples (less than 1\%) from the target domain. Fine-tuning with these carefully selected samples significantly enhances the model's domain adaptability.}
\item{Our proposed DAM-Net has been evaluated on two public datasets, LEVIR-CD and WHU-CD datasets.
Our network achieved beyond state-of-the-art (SOTA) performance, 
and matched the performance of SOTA semi-supervised methods with 10\% of labeled data while only using 0.3\% labeled data.
Additionally, we will release the first open-source implementation of domain adaptation for CD, facilitating future research in this direction.}
\end{enumerate}

The rest of this article is organized as follows.
Section II provides a review of the pertinent literature concerning CD networks and domain adaption.
Section III describes the CD method proposed in this article.
Section IV presents a set of quantitative comparisons and analyses based on experimental results.
Section V presents discussion.
Finally, Section VI concludes this article.

\section{Related Work}

\subsection{CD Networks}

In recent years, neural networks have shown significant promise in geoscience applications, including water quality estimates \cite{hu2023a2dwqpe}, image-text retrieval \cite{he2024visual} and CD \cite{zhu2017change}.
The general architecture of existing deep learning CD networks can be divided into two categories: single-branch networks and multi-branch networks.
Single-branch networks first fuse multi-temporal inputs before feature extraction, so that the image segmentation networks can be conveniently borrowed to solve the CD problem with only slight modifications.
In multi-branch networks, Siamese encoders with shared parameters are typically used to extract distinct feature maps for each temporal input, which are then fused to predict the change map.
Daudt \textit{et al.} \cite{daudt2018urban} built upon the U-Net architecture and introduced early fusion (EF) and Siamese (Siam) strategies, leading to the development of FC-EF, FC-Siam-conc, and FC-Siam-diff \cite{daudt2018fully}, which are widely regarded as the cornerstone of deep learning-based CD networks.
Afterwards, various researchers have made significant contributions to the field of CD networks.
Peng \textit{et al.} \cite{peng2019end} and Fang \textit{et al.} \cite{fang2021snunet} enhanced FC-Siam-conc by developing UNet++\_MSOF and SNUNet-CD, which replaced the U-Net with U-Net++ \cite{zhou2018unet++} and incorporated multiple side outputs from U-Net++.
In a similar vein, Papadomanolaki \textit{et al.} \cite{papadomanolaki2021deep} introduced L-UNet, which integrates fully convolutional LSTM blocks at each encoding stage.
Chen \textit{et al.} \cite{chen2021remote} introduced spatial–temporal attention mechanism and proposed the bitemporal image transformer (BIT), which efficiently captures contextual information in the spatial-temporal domain.
To address specific challenges in CD, several specialized architectures have emerged.
Chen \textit{et al.} \cite{chen2022rdpnet} developed RDP-Net to minimize information loss and enhance detail preservation.
Zhou \textit{et al.} \cite{zhou2023mining} advanced the field with a context aggregation network that leverages inter-image context across training data to enhance intra-image contextual understanding.
Chen \textit{et al.} \cite{chen2024srcnet} have noticed the spatial relationship among multi-temporal inputs, proposed cross-branch interaction modules in the feature extraction stage, paid attention to spatial correlation in the feature fusion stage, and proposed SRC-Net.
\edita{Chang \textit{et al.} \cite{chang2024remote} simplify stacked multi-head transformer blocks and attention mechanism, and propose the multiscale dual-space interactive perception network (MDIP-Net) to Fully utilize ground semantic information.
Huang \textit{et al.} enhance the utilization of multi-scale features, and propose a spatiotemporal enhancement and interlevel fusion network (SEIF-Net) to improve the ability of feature representation for changing objects.}
These existing networks have achieved well performance in CD tasks.
However, when networks trained on the source domain are applied to the target domain, their performance often significantly declines, requiring fine-tuning on the target domain with labeled samples.
This greatly limits the applications of CD networks, which is the central motivation of our work.

\subsection{Domain Adaptation for CD}

Domain Adaptation for image tasks can be generally divided into two categories: one is to translate the target domain image to the source domain, and the other is to reduce the distribution discrepancy of features extracted from the target and source domains.
The method of translating the target domain image to the source domain aims to generate pseudo-images from the target domain that have a style similar to the source domain images, so that predictions trained on the source domain can be directly used for prediction in the target domain.
The representative ones are the Cycle-Consistent Generative Adversarial Network (CycleGAN) proposed by Zhu \textit{et al.} \cite{zhu2017unpaired}, and the CycleGAN-Domain Adaptation (CycleGAN-DA) proposed by Vega \textit{et al.} \cite{vega2021unsupervised}.
The problem with this image-to-image translation method is that pseudo-images translated from the target domain are based on the class distributions of the source domain, which means that land cover information may be incorrectly translated, thereby hindering subsequent predictions.
Meanwhile, the translated images may contain artifacts produced during the translation process, which limits the quality of the translation images.

The method of reducing the distribution discrepancy of features extracted from the target and the source domain is currently the most commonly used in remote sensing \cite{saha2020unsupervised,wittich2021appearance}.
For CD, Chen \textit{et al.} \cite{chen2020dsdanet} using a siamese architecture for spatial-spectral feature extraction, followed by difference vector computation and fully connected layers to generate change detection maps.
Song \textit{et al.} \cite{song2019domain} focused on feature alignment in a common subspace, using a pre-trained CNN and Bregman matrix divergence algorithm.
Elshamli \textit{et al.} \cite{elshamli2017domain} conducted a comparative analysis of DA-based representation learning models, including Denoising Autoencoders.
The above methods are based on feature metric approaches, which aim to represent the features of the source domain and the target domain in the same feature space, thereby minimizing the differences in feature distribution between the source domain and the target domain. These methods heavily rely on finding suitable feature spaces and distance metrics.
With the idea of GAN's Generator-Discriminator, Tsai \textit{et al.} \cite{tsai2018learning} proposed an adversarial learning method for domain adaptation in the context of semantic segmentation. Considering semantic segmentations as structured outputs that contain spatial similarities between the source and target domains, they adopt adversarial learning in the output space.
This brings new ideas to the task of CD.

However, there have been limited studies on domain adaptation for CD in recent years.
In this paper, we introduce ADA into CD, while designing segmentation-discriminators and alternating training strategies according to the characteristics of CD tasks.

\section{Methodology}

In this section, we first briefly introduce the architecture of DAM-Net, followed by our CD-oriented adversarial domain adaptive approach. Finally, we present our micro-labeled fine-tuning to further enhance the network's adaptive capability in the target domain.

\subsection{DAM-Net Architecture}

\begin{figure*}[t]
\centering
\includegraphics[width= \linewidth]{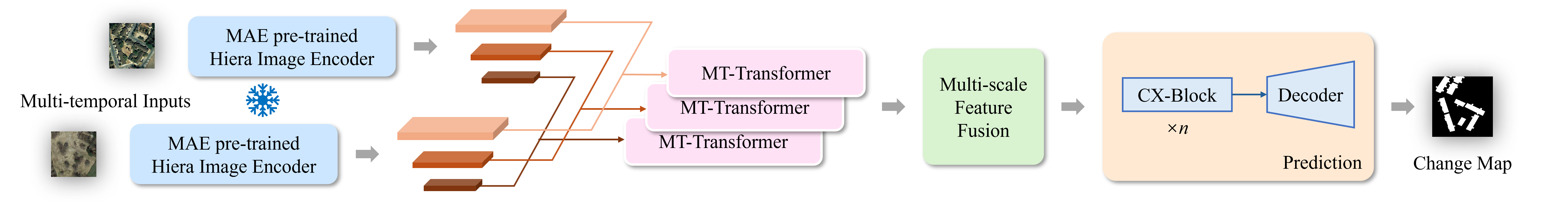}
\caption{Architecture of the proposed DAM-Net.}
\label{DAM}
\end{figure*}

The overall architecture of the proposed DAM-Net is illustrated in Fig. \ref{DAM}.
Given a pair of multi-temporal remote sensing images, the network first extracts multi-scale features through an image encoder. These features are then processed by Multi-Temporal Transformers (MT-Transformers) for temporal fusion, followed by multi-scale feature fusion. Finally, the change detection results are generated through the prediction head.

\textbf{Image Encoder:}
We employ a Hiera \cite{ryali2023hiera} backbone pre-trained with MAE \cite{he2022masked}, which provides hierarchical feature representations enabling effective multi-scale feature extraction.

\begin{figure}[ht]
\centering
\includegraphics[width= 0.6\linewidth]{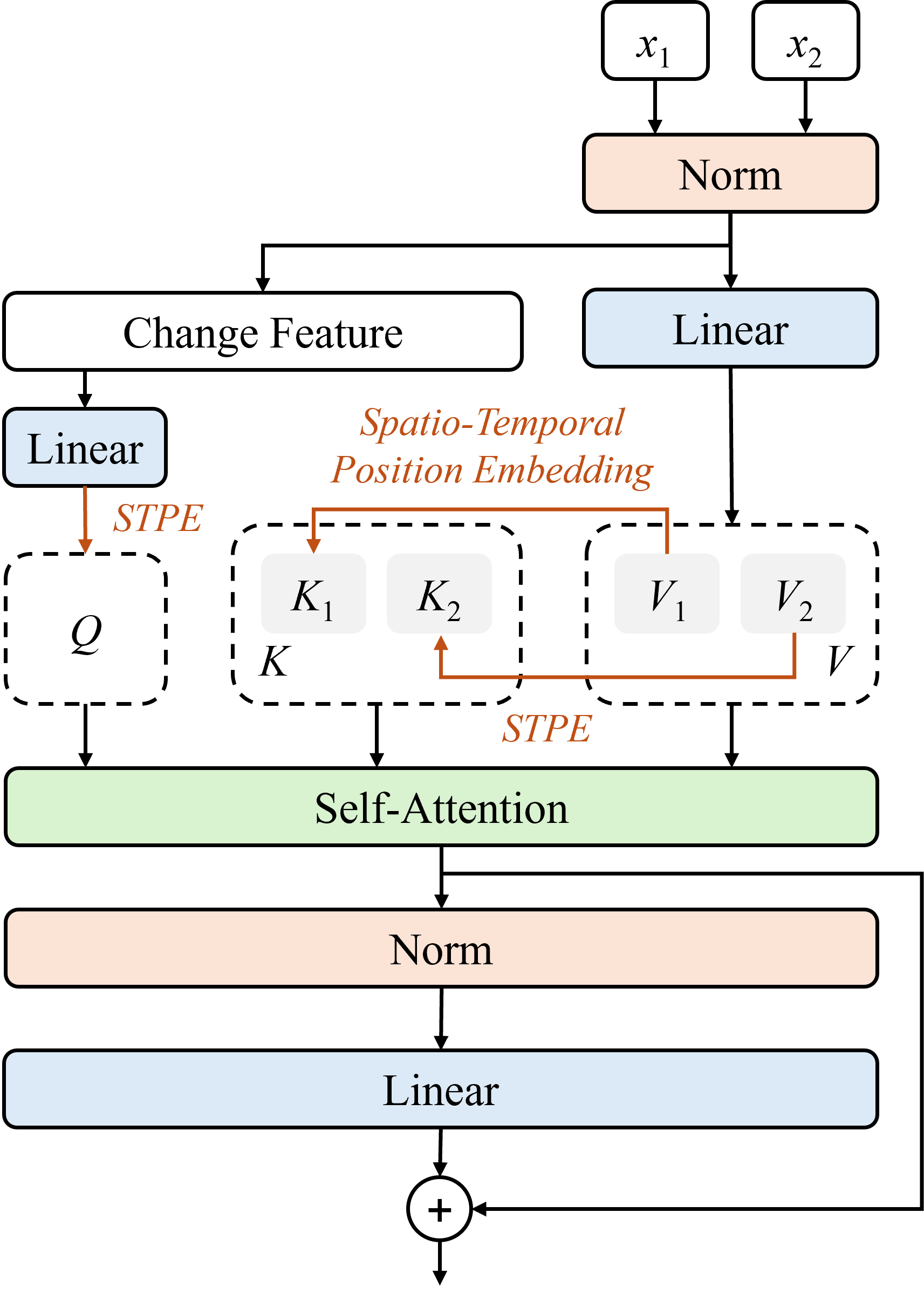}
\caption{Structure of MT-Transformer.}
\label{MT-Transformer}
\end{figure}

\textbf{Multi-temporal Feature Fusion:}
Building upon our previous work on Patch-Mode Feature Fusion Module (PM-FFM) \cite{chen2024srcnet}, we propose the Multi-Temporal Transformer (MT-Transformer) for enhanced temporal feature fusion. As illustrated in Fig. \ref{MT-Transformer}, this module explicitly models temporal relationships to achieve more expressive feature representations.
See Appendix for more details.

\textbf{Multi-scale Feature Fusion:}
We built a multi-scale feature fusion module using a convolutional layer, transposed convolutional layers, normalization layers and activation layers to fuse multi-scale features encoded by the image encoder.

\textbf{Prediction Head:}
The prediction head consists of multiple CX-Blocks \cite{woo2023convnext} followed by a decoder, which processes the fused multi-scale features to generate the final change detection map.

The source code of DAM-Net will be made publicly available at \url{https://github.com/chnja} upon publication.

\subsection{Adversarial Domain Adaptation for CD}
\label{secada}

ADA was first introduced to the semantic segmentation task by Tsai \textit{et al.} \cite{tsai2018learning}.
Their approach consists of a segmentation network and multi-level discriminators, processing both labeled source domain and unlabeled target domain images. The segmentation network first learns from source images, then generates predictions for both domains, while discriminators identify the domain origin of these predictions.
Through adversarial training, the segmentation network learns to generate similar prediction distributions across domains, enabling adaptation to the unlabeled target domain.
During training, their approach designed a hybrid loss to jointly train both the segmentation network and the domain discriminator.

We introduce the above ADA framework into CD and propose three key improvements based on CD characteristics:
Refined CD-oriented segmentation-discriminator architecture, improved discriminator output design, and alternating training strategy. These improvements collectively enhance the domain adaptation capability for CD tasks.

\begin{figure*}[ht]
\centering
\includegraphics[width= 0.8\linewidth]{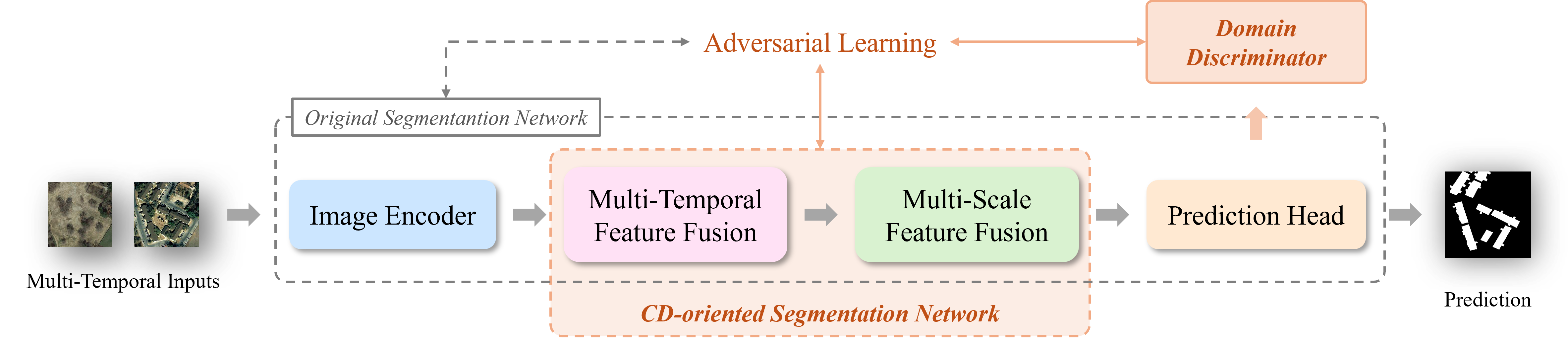}
\caption{Simplified architecture of our ADA.}
\label{simple}
\end{figure*}

\textbf{Segmentation-Discriminator:}
Unlike the original ADA approach that the entire network was treated as a segmentation network, with all parameters being updated during training, we propose a more efficient architecture (Fig. \ref{simple}).
Our network comprises four key modules: image encoder, multi-temporal feature fusion, multi-scale feature fusion, and prediction head.
This is also a common architecture for CD networks.
The image encoder is primarily responsible for extracting features from input images, and increasingly, researchers opt to utilize pre-trained modules trained on large-scale datasets for this purpose.
Multi-temporal feature fusion focuses on fusing and describing changes between multi-temporal features, making it crucial for CD tasks.
Multi-scale feature fusion combines previously extracted multi-scale features, which is particularly significant for domain adaptation due to variations in target object sizes across different datasets.
The prediction head generates final results using the extracted feature maps.

We argue that for the CD task, even across different datasets, the main change feature extraction has already been completed by the first three stages, and the prediction head parameters can be generalizable. Moreover, the image encoder has been sufficiently trained on large-scale datasets, so it requires minimal or no fine-tuning.
During domain adaptation, only the parameters of multi-temporal and multi-scale feature fusion need to be updated, which reduces computation and improves training speed.
The corresponding experiments in section \ref{expada} have proved our viewpoint.

Furthermore, the discriminator's input was the final segmentation result in the original algorithm, primarily because it was designed for images captured by forward-facing cameras in vehicles.
Such design is feasible when the size, position, and quantity of segmentation results in the source and target domains are not significantly different.
However, for CD, there may be substantial differences in the size, position, and quantity of change targets across domains. If segmentation results are input into the discriminator, even with good segmentation performance, the discriminator could easily determine the input domain, hindering the adversarial learning process in adversarial learning.
Therefore, we use intermediate results from the prediction head as discriminator inputs, and modify the discrimination of the prediction result to the discrimination of the extracted features.
This improvement facilitates better feature space alignment and domain adaptation for CD.

\textbf{Matrix-Output Discriminator:}
In the original algorithm and GANs, discriminators have always output a single classification result for single input.
This was reasonable for previous tasks where discriminators' domain judgment came from perceiving the entire input image.
Taking semantic segmentation of vehicle-captured images as an example: roads typically appear at the bottom, and the sky at the top.
The discriminators' judgment targeted the entire input image.
However, in CD tasks, change areas have no fixed position.
Even inputting part of the feature map into the discriminator should enable domain discrimination.
Therefore, we modified the domain discriminator to output a matrix,
thereby enabling domain discrimination for different regions on the input feature map.
This improvement enables the segmentation network to better align the feature distributions between domains, while allowing the segmentation and discriminator to focus more on poorly-adapted and well-discriminated regions, thereby enhancing domain adaptation performance.

\textbf{Alternating Training Strategy:}
In GANs, the alternating training strategy usually involves alternate training of the discriminator and generator.
First, one batch trains the discriminator to distinguish generator results, then the next batch trains the generator to deceive the discriminator, cycling continuously.
While in the ADA algorithm, the training is primarily divided into three parts:
training segmentation network on the labeled source domain, training the discriminator to distinguish the domain of extracted feature maps, and training the segmentation to deceive the discriminator.
However, it uses a hybrid loss combining segmentation loss on source domain and adversarial loss, with gradient reversal applied to the discriminator, rather than alternating training.
The weights of these losses are manually set.
In CD tasks, we found that loss weight selection is particularly crucial, as the network can easily prioritize deceiving the discriminator while compromising CD performance.

\begin{table*}[ht]
\caption{Steps of Alternating Training Strategy.}
\label{tab:ats}
\centering
\begin{tabular}{cccc}
\hline
Step & Description & Training Domain & Training Parts\\
\hline
1 & Training DAM-Net & Source domain & Multi-Temporal Feature Fusion, Multi-Scale Feature Fusion, Prediction\\
2 & Training Discriminator & Source domain, target domain & Discriminator\\
3 & Training segmentation network & Source domain, target domain & Multi-Temporal Feature Fusion, Multi-Scale Feature Fusion\\
\hline
\end{tabular}
\end{table*}

Therefore, we redesigned the alternating training strategy, considering our segmentation-discrimination improvements, as detailed in Table \ref{tab:ats}.
The training comprises three steps:
First, DAM-Net are trained in the source domain to prevent excessive focus on domain adaptation at the expense of CD performance.
Second, the domain discriminator is trained using prediction feature maps from both domains to distinguish input domains.
Third, the segmentation is trained using prediction feature maps from both domains to deceive the discriminator into believing all inputs are from the source domain.
The training loops through these steps continuously.
In this adversarial learning process, domain adaptation for CD is accomplished.
During training, the first step uses a hybrid loss fuction from \cite{chen2024srcnet}, while the second and third steps use binary cross entropy loss function.

With these three improvements, we successfully adapt the ADA framework to CD tasks by adjusting the segmentation-discriminator structure and redesigning the alternating training strategy.
These improvements reduce computational overhead and increase training speed while enhancing robust CD performance.

\subsection{Micro-Labeled Fine-Tuning (MLFT)}

In Adversarial Domain Adaptation (ADA), while our approach shows promising results, the complete absence of target domain labels may limit potential performance improvements.
This raises an intriguing question: \textit{Can we achieve better adaptation performance by utilizing just a minimal amount of labeled data from the target domain?}
Furthermore, we argue that not all samples contribute equally to the domain adaptation process. Some samples are more challenging to adapt than others, and these challenging cases often represent the key of domain adaptation.
Based on these observations, we propose a novel Micro-Labeled Fine-Tuning approach that selects key samples in domain adaptation for fine-tuning, thereby better completing domain adaptation under extremely limited labeled sample conditions.

The first step is to select suitable key samples, MLFT leverages the output of domain discriminator, which has been well-trained in ADA, to intelligently identify the most valuable samples for annotation, instead of random selection or manual identification of samples.
The domain discriminator, trained to determine whether features originate from source or target domains, provides a unique perspective on domain adaptation effectiveness.
Samples that the discriminator more confidently classifies as target domain indicate less effective domain adaptation. Conversely, samples with source domain classification suggest more successful feature alignment.
By selecting samples from high to low probability of coming from the target domain based on domain the discriminator, we strategically choose samples which are more beneficial for further enhancing the network's performance.
During the selection process, we intentionally avoid selecting samples with minimal change regions according to the detection results, arguing that such samples provide limited adaptation insights and might cause the network to focus on irrelevant details.
Typically, our approach requires annotating only 10-20 samples in the target domain to achieve significant performance improvements.
Here, we use the cross entropy loss function ${\mathcal L}_{ML}$.

\begin{figure}[ht]
\centering
\includegraphics[width= 0.5\linewidth]{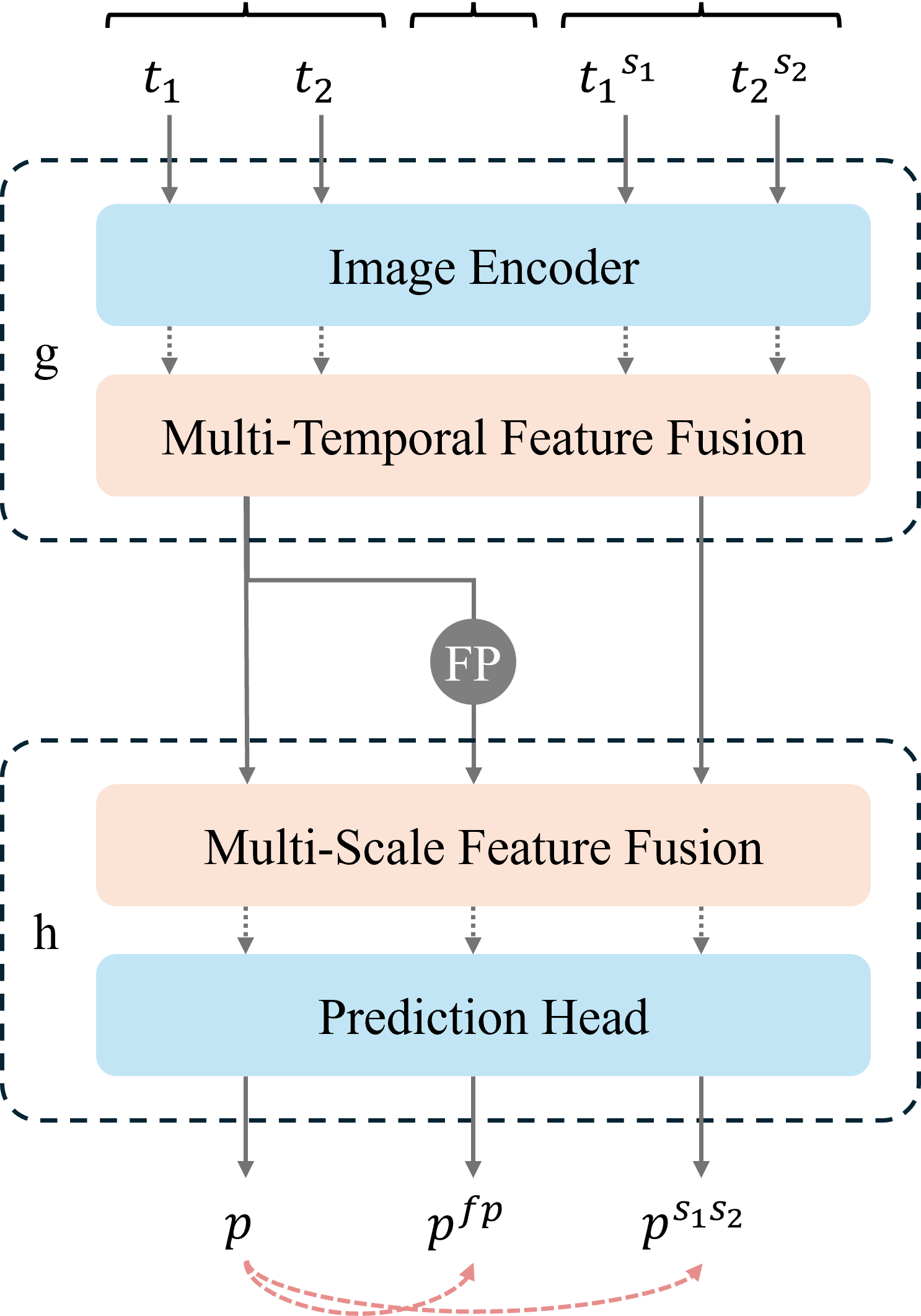}
\caption{Framework of CDMatch.}
\label{MLFT}
\end{figure}

However, if we train the network using only these labeled samples, the network is very prone to overfitting.
Therefore, we also need to generate pseudo-labels for a large amount of unlabeled data.
At this point, we introduced the idea of FixMatch \cite{sohn2020fixmatch} and UniMatch \cite{yang2023revisiting}, and proposed CDMatch.
The framework of CDMatch is shown in Fig. \ref{MLFT}.
We apply strong image perturbations to the network input and feature perturbations to the extracted features. And the network would be supervised through the consistency loss between the unperturbed and perturbed outputs, thereby enhancing the robustness of the network's feature extraction.
The CDMatch framework applies perturbation strategies to enhance network robustness:
1) Image Perturbations: Apply strong image perturbations to multi-temporal inputs respectively.
2) Feature Perturbations: Apply feature perturbations to extracted feature maps.
This process can be described as
\begin{equation}
\label{s1234}
\begin{split}
t_1^{s_1}=S_1\left(t_1\right), t_2^{s_2}=S_2\left(t_2\right)
\end{split}
\end{equation}
\begin{equation}
\begin{split}
&p=h\left(g\left(t_1, t_2\right)\right)\\
&p^{fp}=h\left(FP\left(g\left(t_1, t_2\right)\right)\right)\\
&p^{s_1s_2}=h\left(g\left(t_1^{s_1}, t_2^{s_2}\right)\right)
\end{split}
\end{equation}
where $t_1, t_2$ represent the original multi-temporal inputs, $S_1(), S_2()$ represent four strong image perturbations. The Eq. \ref{s1234} represents applying disturbances $S_1$ to $t_1$ and $S_2$ to $t_2$, resulting in images $t_1^{s_1}, t_2^{s_2}$.
$g()$ represents image encoder and multi-temporal feature fusion, $h()$ represents multi-scale feature fusion and prediction head, $FP()$ represent the feature perturbation.
$p$ represents the prediction result of the original input $t1, t2$, $p^{fp}$ represents the prediction results after feature perturbation, $p^{s_1s_2}$ represents the prediction results after image perturbation.
We use $p$ as pseudo-labels and calculate the difference between $p, p^{fp}, p^{s_1s_2}$ using binary cross entropy loss $BCE()$. The loss function of CDMatch can be fomulated as ${\mathcal L}_{CM}$:
\begin{equation}
{\mathcal L}_{CM}=\left(BCE\left(p,p^{fp}\right)+BCE\left(p,p^{s_1s_2}\right)\right)/2
\end{equation}
Besides, to ensure the network's performance, we continued to utilize source domain data during fine-tuning, with the cross entropy loss function denoted as ${\mathcal L}_{S}$.
The loss at this stage can be fomulated as ${\mathcal L}_{FT}$:
\begin{equation}
{\mathcal L}_{FT}=\alpha{\mathcal L}_{CM}+\beta{\mathcal L}_{S}+\gamma{\mathcal L}_{ML}
\end{equation}
where $\alpha, \beta, \gamma> 0$ are coefficients.

Through this innovative approach, we successfully prevented network overfitting while training with very few labeled samples, significantly enhanced the network's performance in the target domain, and achieved robust domain adaptation. By integrating intelligent sample selection and fine-tuning strategies, our method provides a flexible framework that minimizes manual annotation requirements while bridging the gap between different domains.

\section{Experiment and Result Analysis}

\subsection{Dataset}

The experiment was conducted on two datasets named LEVIR-CD dataset (LEVIR-CD) \cite{chen2020spatial} and WHU Building dataset (WHU-CD) \cite{ji2018fully}, two of the most common datasets in remote sensing CD.

\textit{LEVIR-CD} dataset consists of 637 visible light image pairs of $1024\times 1024$ pixels, which are collected from Google Earth. The spatial resolution is 0.5m per pixel. We cut each image pair into $256\times 256$ pixel patches without overlapping and ultimately obtained 7120 training sets and 1024 validation sets.

\textit{WHU Building} dataset consists of two visible light aerial images of size $32507\times 15354$, which are captured by NZ Aerial Mapping Ltd. The spatial resolution is 0.3 m per pixel. We cut each image pair into $256\times 256$ pixel patches without overlapping and ultimately obtained 5141 training sets and 2293 validation sets.

\subsection{Implementation Details}

We implemented DAM-Net using the PyTorch framework. The number of CX-Blocks in the prediction head is set to 3, and we employed AdamW \cite{loshchilov2017decoupled} as the optimizer. 
In ADA stage, the learning rate is set to 1e-5 and decays by 0.8 every 5 epochs.
The training process consisted of three steps with different batch sizes: the first step used a batch size of 16, while the second and third steps employed a batch size of 32, with 16 samples each from source and target domain.
During MLFT stage, the learning rate is set to 5e-6 and decays by 0.8 every 5 epochs. The batch size is 61, with 15 unlabeled target domain samples, 30 perturbations generated by CDMatch, 1 labeled target domain fine-tuning sample, and 15 labeled source domain samples.
The corresponding weight coefficients $\alpha, \beta, \gamma$ are set to $30/46, 1/46, 15/46$, respectively.
All experiments were conducted on either a single NVIDIA RTX3090 or Tesla A100 GPU for 100 epochs.

We used five standard metrics, precision, recall, F1 score, overall accuracy (OA), and intersection over union (IoU) to quantitatively evaluate the performance of the CD models.

\subsection{Comparison With SOTA Methods}

Domain adaptation for CD remains a relatively unexplored field. Notable contributions include CGDA-CD by Vega \textit{et al.} \cite{vega2021unsupervised} and SFDA-CD by Wang \textit{et al.} \cite{wang2024sfda}, which represent pioneering efforts in this direction. CGDA-CD, the first attempt at unsupervised DA for CD, adapts the CycleGAN model for image pair translation and introduces novel regularization constraints. SFDA-CD advances the field by proposing a source-free unsupervised DA architecture that operates solely with a pretrained source model and unlabeled target data.

However, these approaches have certain limitations.
CGDA-CD treats the cross-domain CD task as separate components: it independently transfers pre-temporal and post-temporal images from the target domain to the source domain, followed by change prediction using a source-domain-trained CD network. This segmented approach fails to address domain adaptation for CD holistically. Moreover, the absence of publicly available implementation code hampers further research and verification.
While SFDA-CD represents the first framework to consider CD domain adaptation as an integrated task, it also lacks open-source code and sufficient implementation details for reproducibility. Given these constraints, we can only cite experimental results from relevant papers.
Through our analysis, we observed that the LEVIR-CD dataset contains more diverse and richer scenarios compared to the WHU-CD dataset. As CGDA-CD and SFDA-CD are designed for special domain adaptation tasks in CD, with some of their modules pre-trained across all CD datasets \cite{vega2021unsupervised, wang2024sfda}, it would be inappropriate to directly compare their detection performance with other domain adaptation methods. Consequently, we focus our comparison on the LEVIR-CD $\rightarrow$ WHU-CD transfer scenario.
For the MLFT component, which involves labeled samples, we present comparisons with state-of-the-art semi-supervised CD methods in Section \ref{lMLFT}.
To promote advancement in CD domain adaptation research, we will release the complete implementation of DAM-Net, including source code and documents.

\begin{table*}[ht]
\caption{Results on LEVIR-CD $\rightarrow$ WHU-CD and WHU-CD $\rightarrow$ LEVIR-CD.}
\label{tab:result}
\centering
\begin{threeparttable}
\begin{tabular}{ccccccccccc}
\hline
\hline
\multirow{2}*{Model} & \multicolumn{5}{c}{LEVIR-CD $\rightarrow$ WHU-CD} & \multicolumn{5}{c}{WHU-CD $\rightarrow$ LEVIR-CD}\\
\cline{2-11}
 & Precision & Recall & F1 & OA & IoU & Precision & Recall & F1 & OA & IoU\\
\hline
CGDA-CD\dag & 0.7119 & 0.5622 & 0.6283 & - & 0.5233 & - & - & - & - & -\\
SFDA-CD\dag & 0.7372 & 0.5617 & 0.6376 & - & 0.5377 & - & - & - & - & -\\
\hline
DAM-Net w/o MLFT & 0.8090 & 0.6827 & 0.7405 & 0.9788 & 0.5879 & 0.7037 & 0.5296 & 0.6044 & 0.9709 & 0.4330\\
DAM-Net (16) & 0.8751 & 0.7911 & 0.8309 & 0.9860 & 0.7108 & 0.6916 & 0.6983 & 0.6949 & 0.9645 & 0.5325\\
\hline
\hline
\end{tabular}
\begin{tablenotes}
\footnotesize
\item[\dag] Source code is not publicly available, results are directly cited from the research paper.
\end{tablenotes}
\end{threeparttable}
\end{table*}

Table \ref{tab:result} reports the comparisons of detection accuracy.
Our proposed DAM-Net (without micro-labeled fine-tuning) outperforms existing SOTA methods across all key metrics (precision, recall, F1, and IoU).
For the LEVIR-CD $\rightarrow$ WHU-CD adaption task, DAM-Net achieves substantial improvements over the current SOTA: precision increased by 7.18\%, recall by 12.05\%, F1 by 10.29\%, and IoU by 5.02\%.
The integration of micro-labeled fine-tuning further enhances performance significantly. With merely 16 labeled samples, we observe additional gains in all metrics: precision (+6.61\%), recall (+10.84\%), F1 (+9.04\%), OA (+0.72\%), and IoU (+12.29\%).
For the WHU-CD $\rightarrow$ LEVIR-CD adaption task, despite the limited scenario diversity in the WHU-CD dataset, DAM-Net achieves promising results with 70.37\% precision and 60.44\% F1-score.
The effectiveness of our MLFT strategy is further validated in this direction, where fine-tuning with just 16 labeled samples yields substantial improvements in both F1 (+9.05\%) and IoU (+9.95\%).

\begin{figure*}[!ht]
\centering
\subfloat[]{
    \includegraphics[width= 0.16\textwidth]{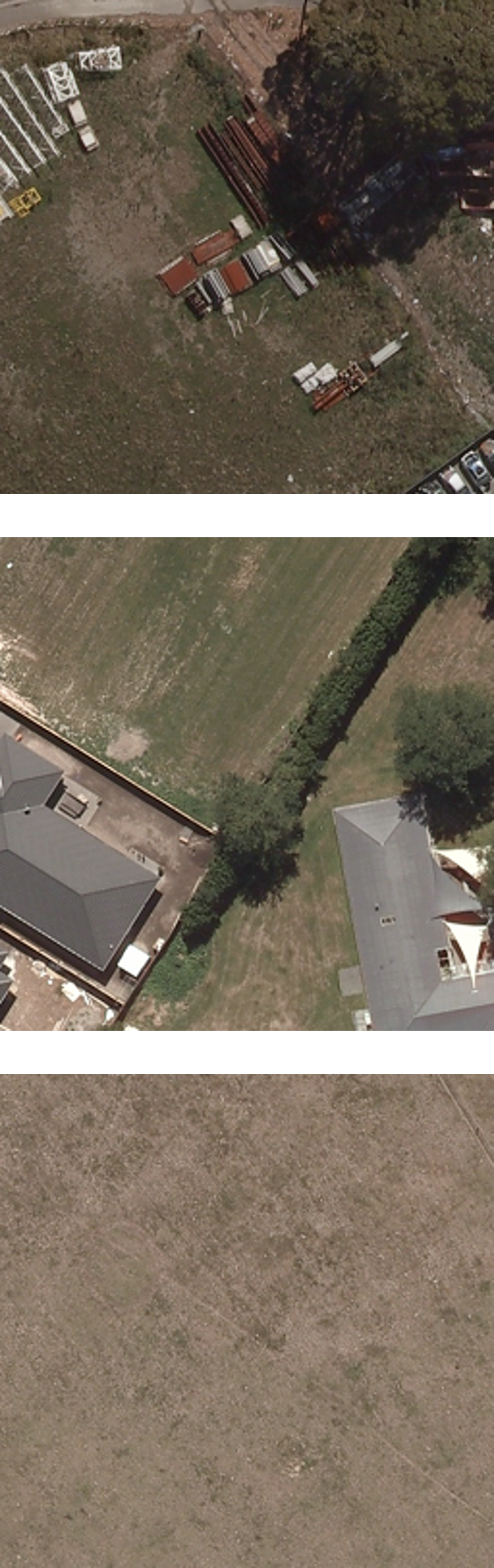}
}
\subfloat[]{
    \includegraphics[width= 0.16\textwidth]{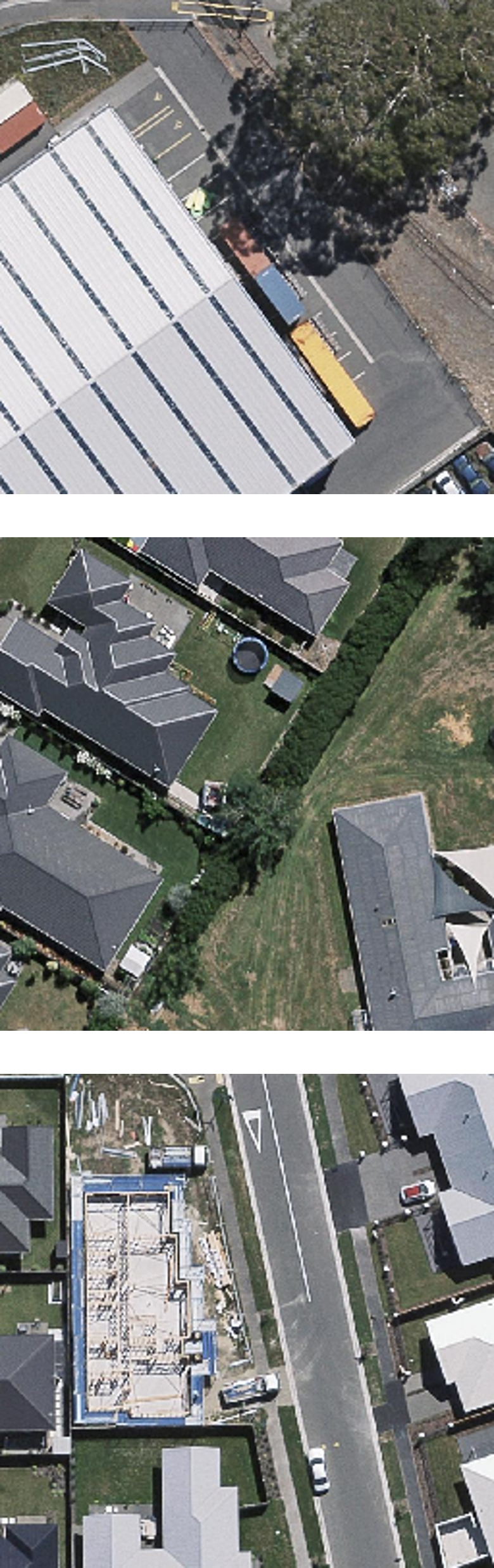}
}
\subfloat[]{
    \includegraphics[width= 0.16\textwidth]{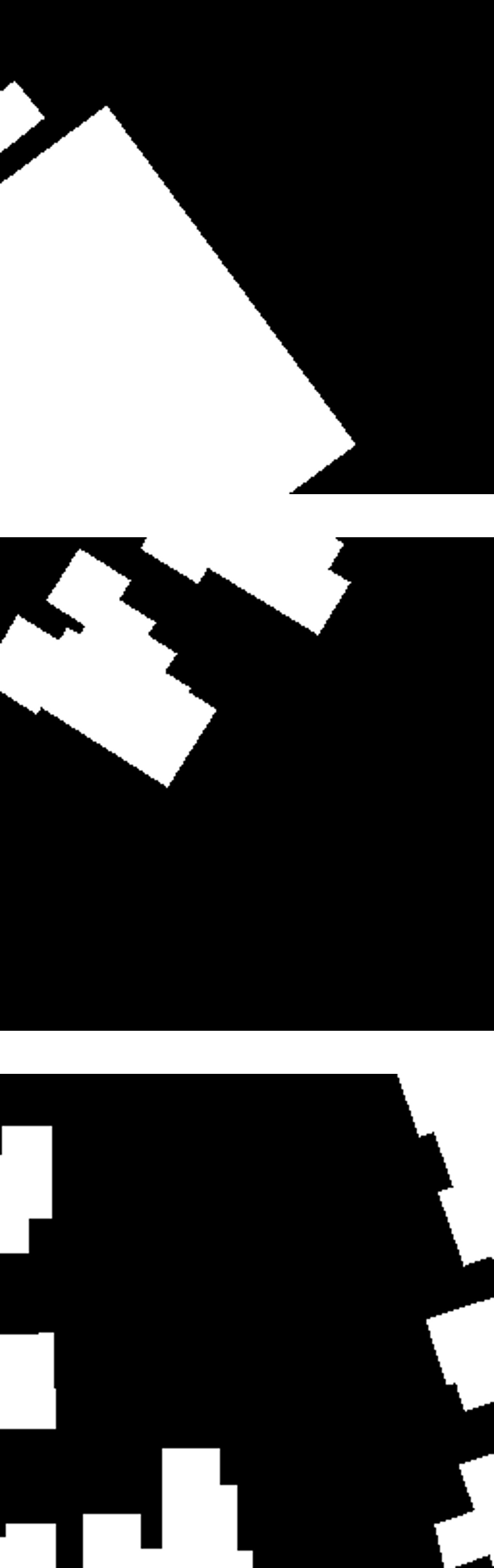}
}
\subfloat[]{
    \includegraphics[width= 0.16\textwidth]{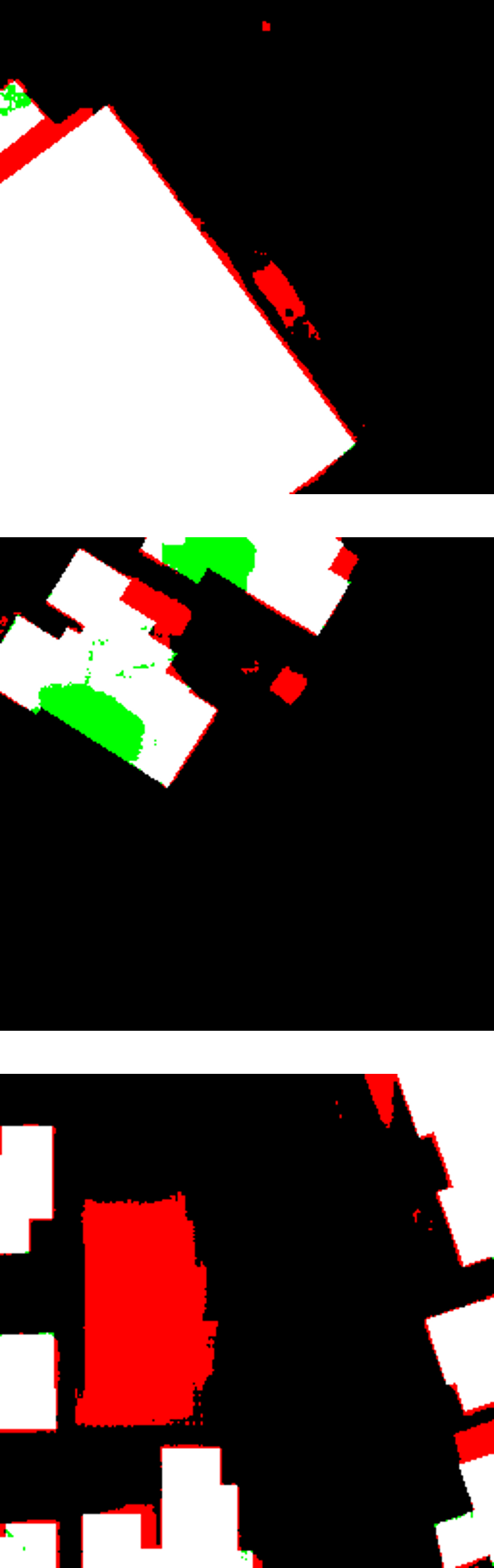}
}
\subfloat[]{
    \includegraphics[width= 0.16\textwidth]{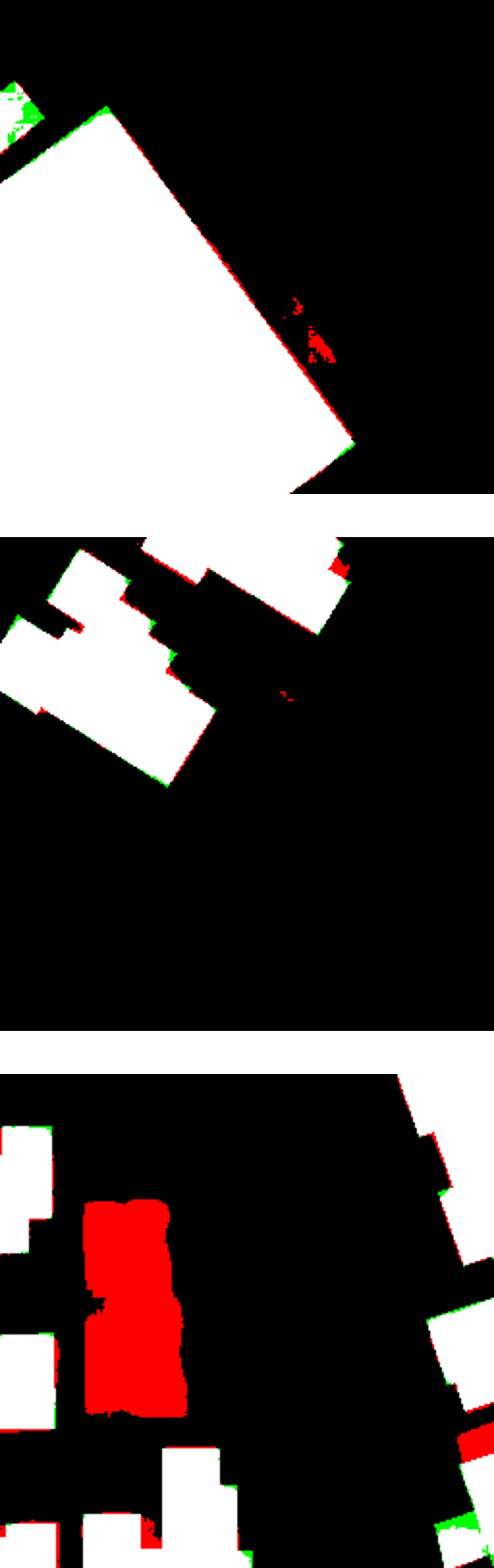}
}

\subfloat[]{
    \includegraphics[width= 0.16\textwidth]{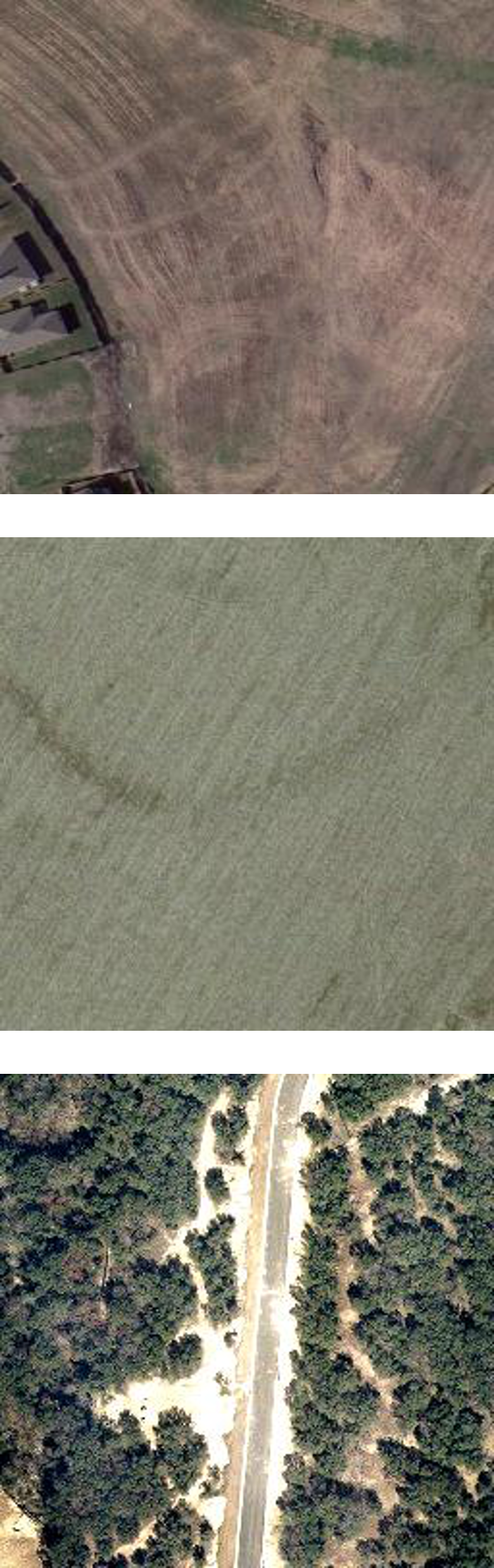}
}
\subfloat[]{
    \includegraphics[width= 0.16\textwidth]{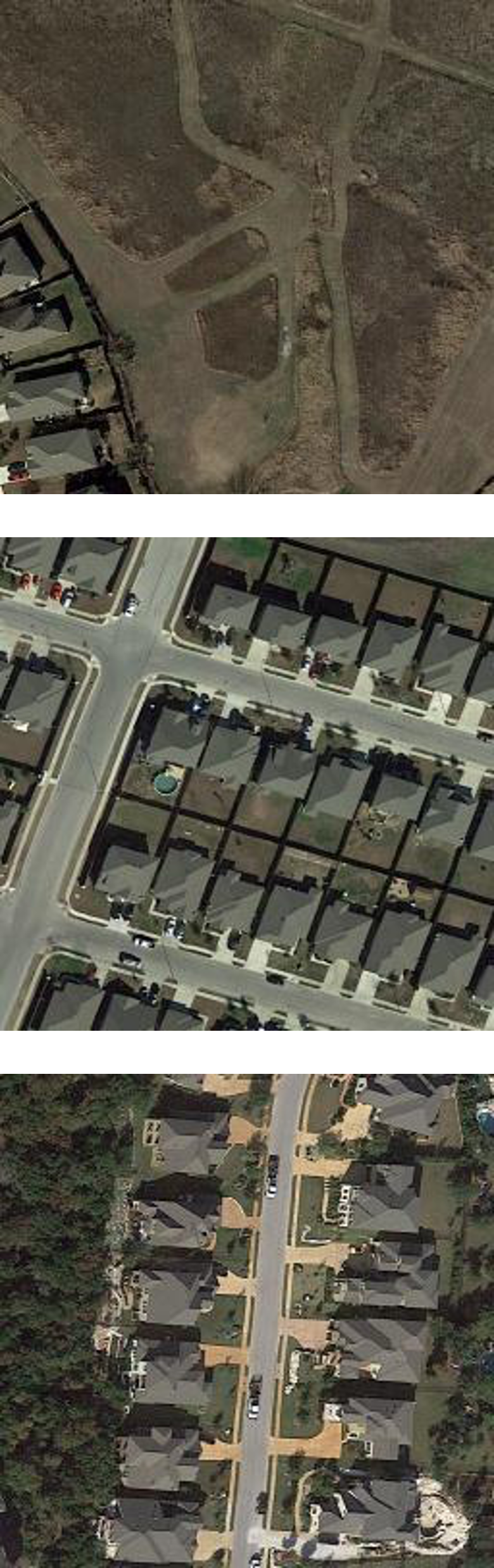}
}
\subfloat[]{
    \includegraphics[width= 0.16\textwidth]{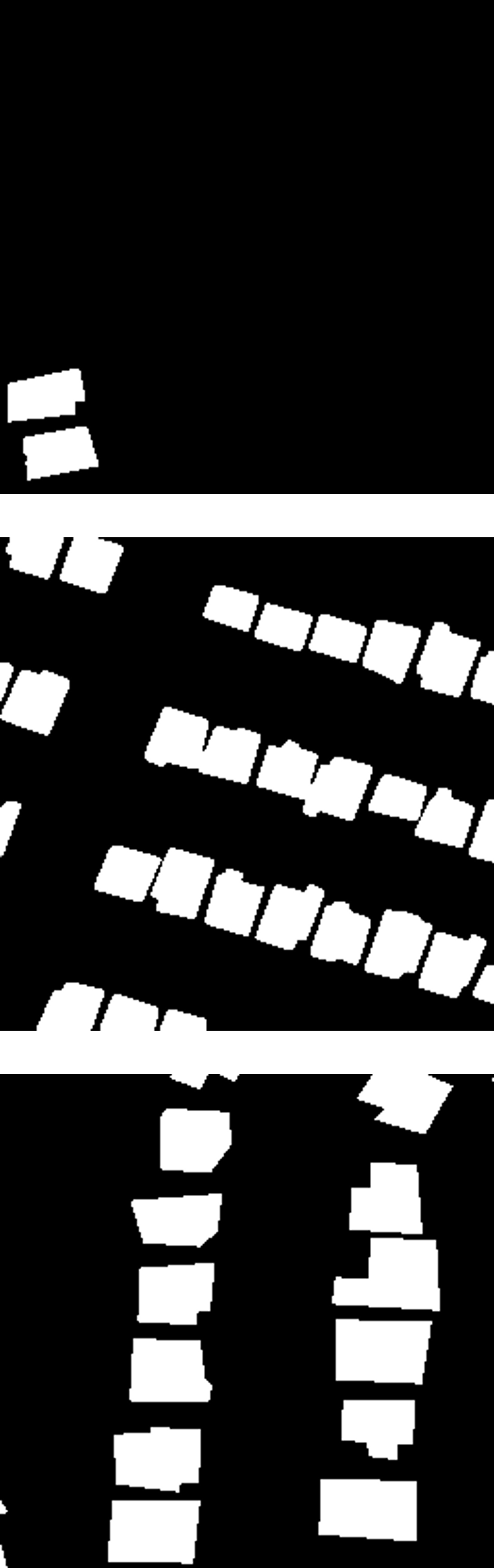}
}
\subfloat[]{
    \includegraphics[width= 0.16\textwidth]{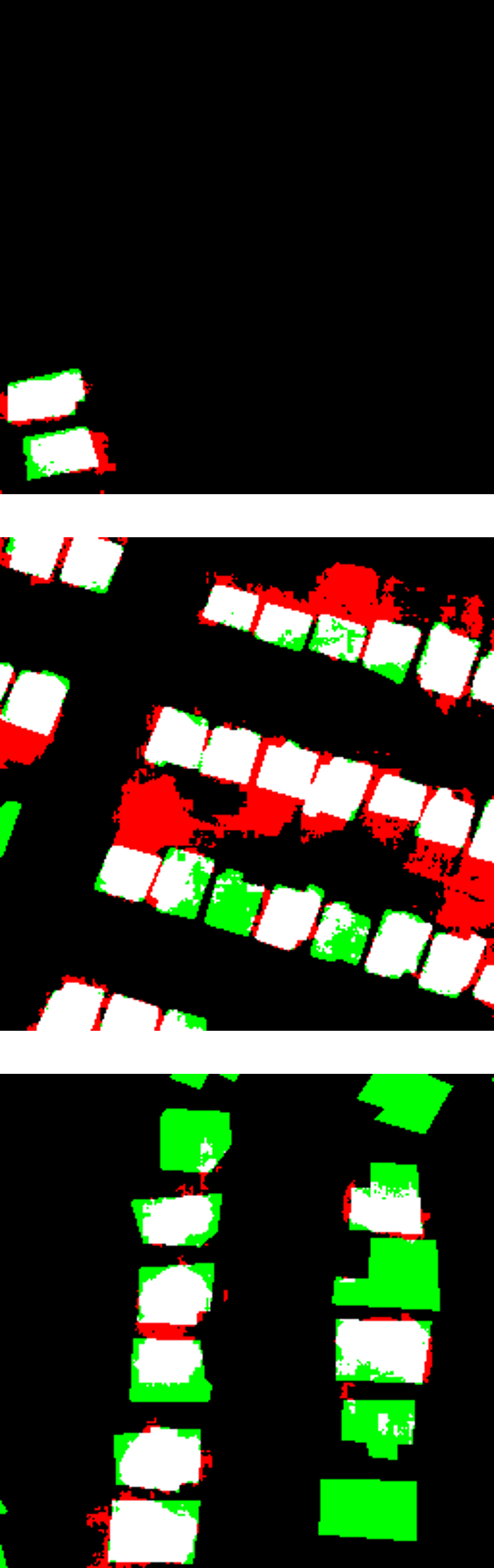}
}
\subfloat[]{
    \includegraphics[width= 0.16\textwidth]{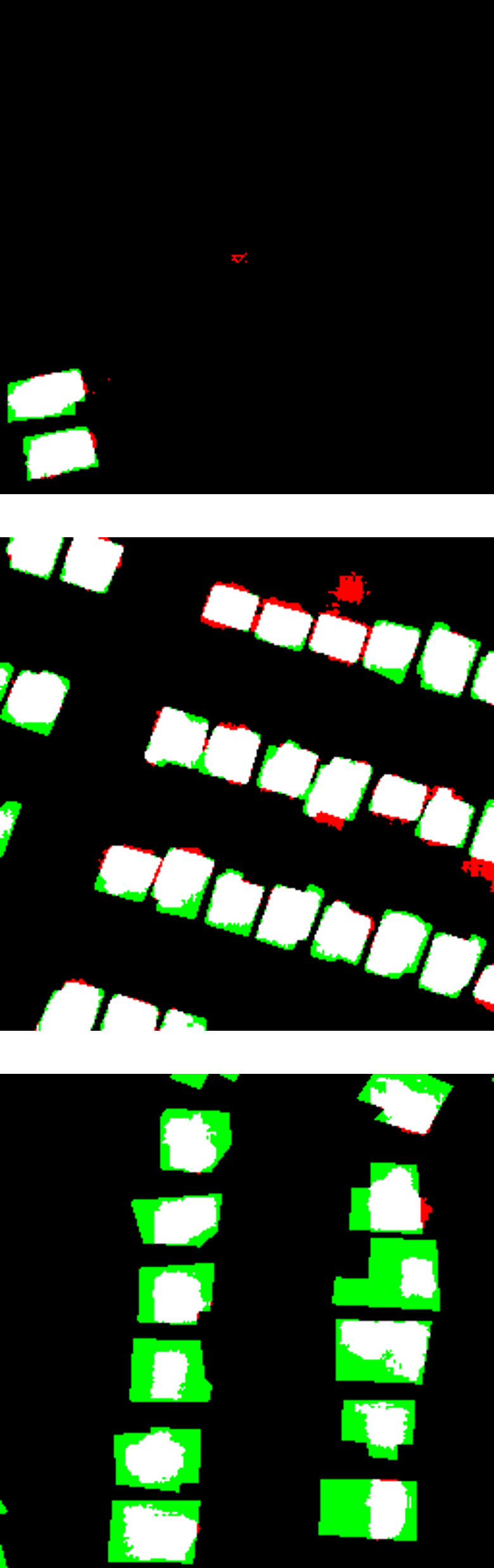}
}
\caption{(a)-(e) are results from LEVIR-CD $\rightarrow$ WHU-CD.
(f)-(j) are results from WHU-CD $\rightarrow$ LEVIR-CD.
(a), (b), (f), and (g) are the original images.
(c) and (h) are the ground truth.
The results of (d) (i) our DAM-Net w/o MLFT,
(e) (j) our DAM-Net.
The false positives and false negatives are indicated in red and green, respectively.
Other colors represent true positives.}
\label{SOTA}
\end{figure*}

In addition, Fig. \ref{SOTA} illustrates representative detection results for both adaption directions (LEVIR-CD $\rightarrow$ WHU-CD and WHU-CD $\rightarrow$ LEVIR-CD).
The false positives and false negatives are indicated by red and green, respectively. Other colors represent true positives.
We present three characteristic scenarios from the validation set: (1) cases where DAM-Net (without MLFT) achieved strong performance through ADA, (2) cases where MLFT significantly improved detection accuracy, and (3) challenging cases with suboptimal performance.
Through detailed analysis, we found that samples with poor performance primarily represented scenarios not present in their respective source datasets, creating inherent difficulties for domain adaptation.
In the LEVIR-CD $\rightarrow$ WHU-CD adaption task, 54.2\% of samples achieved F1 exceeding the baseline ADA performance (0.7405).
MLFT improved F1 scores in 64.8\% of cases, with 65.9\% of these improvements exceeding 5\%.
14.3\% of samples remained challenging even after MLFT
After MLFT, only 14.3\% of samples had F1 lower than the original overall F1 (0.7405).
As shown in the example in the figure, the challenging detection area is a construction site, which is rarely found in LEVIR-CD dataset.
In the WHU-CD $\rightarrow$ LEVIR-CD adaption task, 48.0\% of samples achieved F1 exceeding the baseline ADA performance (0.7405).
MLFT improved F1 scores in 63.5\% of cases, with 78.9\% of these improvements exceeding 5\%.
26.3\% of samples remained challenging even after MLFT, primarily due to scenario types (dense tree scenes, etc.) absent from the source dataset (WHU-CD).

The consistent performance improvements across both adaption directions demonstrate the robustness and generalizability of our approach. These results underscore DAM-Net's significant contribution to advancing domain adaptation research in CD tasks, particularly in scenarios with limited labeled data availability.

\subsection{Experiment in Adversarial Domain Adaptation}
\label{expada}

This part of the experiments focus on the three improvements we made when applying the domain adaptation algorithm to the CD.

First, we conducted comprehensive ablation experiments to evaluate our idea regarding the segmentation-discriminator architecture. 
We argued that for ADA, it would be more effective to train only the multi-temporal feature fusion and multi-scale temporal feature fusion components while keeping the prediction head frozen.
This approach not only promises better adaptation performance but also offers computational advantages through reduced parameter optimization.
To demonstrate this, we experimented with various combinations of training and freezing structures to compare their performance, as shown in Table \ref{tab:ada}, where the checkmarks indicate which parts participated in training:
\begin{itemize}
\item{$a100$: Training Multi-Temporal Feature Fusion}
\item{$a010$: Training Multi-Scale Feature Fusion}
\item{$a001$: Training Prediction Head}
\item{$a111$: Training Multi-Temporal Feature Fusion, Multi-Scale Feature Fusion, and Prediction Head}
\item{$a110$: Training Multi-Temporal Feature Fusion, and Multi-Scale Feature Fusion (DAM-Net)}
\end{itemize}
From the results of the experiment, when training the three parts separately, $a100$ shows better performance (F1: 0.7365, IoU: 0.5829) than $a010$ (F1: 0.7277, IoU: 0.5720), which in turn performs better than $a001$ (F1: 0.7031, IoU: 0.5422).
The performance of training Multi-Temporal Feature Fusion alone $a100$ (F1: 0.7365, IoU: 0.5829) even surpasses training all three parts simultaneously $a111$ (F1: 0.7351, IoU: 0.5811).
Meanwhile, the second-best evaluation metrics also primarily comes from $a100$ and $a010$.
By freezing the Prediction Head and only training the Multi-Temporal Feature Fusion and Multi-Scale Feature Fusion parts ($a110$), we achieved the best overall performance with F1 of 0.7405, OA of 0.9788, and IoU of 0.5879, surpassing the full model ($a111$) across all major metrics.
This improvement not only enhanced the network's performance but also reduced the trainable parameters significantly, resulting in faster training convergence compared to training all components ($a111$).
As analyzed in section \ref{secada}, the Prediction Head part has already been sufficiently trained on the source domain. Continuing fine-tuning during domain adaptation would increase the search space for optimal network parameters and make network training more challenging.
These results validate our idea that focusing the domain adaptation on temporal and scale feature fusion components while freezing the pre-trained prediction head leads to more effective and efficient adaptation, as evidenced by improved performance metrics and reduced computational overhead.

\begin{table*}[ht]
\caption{Results with Adversarial Domain Adaptation on LEVIR-CD $\rightarrow$ WHU-CD.}
\label{tab:ada}
\centering
\begin{threeparttable}
\begin{tabular}{ccccccccc}
\hline
\hline
Method & Multi-Temporal Fusion & Multi-Scale Fusion & Prediction Head & Precision & Recall & F1 & OA & IoU\\
\hline
$a100$ & \ding{52} & & & 0.7882 & \textbf{0.6912} & \underline{0.7365} & 0.9780 & \underline{0.5829} \\
$a010$ & & \ding{52} & & \underline{0.8225} & 0.6526 & 0.7277 & \underline{0.9782} & 0.5720 \\
$a001$ & & & \ding{52} & \textbf{0.8372} & 0.6061 & 0.7031 & 0.9773 & 0.5422 \\
$a111$ & \ding{52} & \ding{52} & \ding{52} & 0.7967 & 0.6823 & 0.7351 & 0.9782 & 0.5811 \\
$a110$ & \ding{52} & \ding{52} & & 0.8090 & \underline{0.6827} & \textbf{0.7405} & \textbf{0.9788} & \textbf{0.5879} \\
\hline
\hline
\end{tabular}
\begin{tablenotes}
\footnotesize
\item The first-best value is marked in bold, and the second-best value is underlined.
\end{tablenotes}
\end{threeparttable}
\end{table*}

Second, we conducted experiments on our improved discriminator design.
Unlike conventional discriminators that output a single discrimination result for the entire input, our approach produces a discrimination matrix that provides pixel-wise adversarial supervision. To validate this improvement, we performed comparative experiments between:
\begin{itemize}
\item{$b1$: Conventional discriminator}
\item{$b2$: Proposed matrix-output discriminator (DAM-Net)}
\end{itemize}
The experimental results in Table \ref{tab:ada2} demonstrate that our matrix-output discriminator consistently outperforms the conventional discriminator across all evaluation metrics. Specifically, compared to $b1$, our improved discriminator ($b2$) achieved higher precision (80.90\% vs 80.73\%), recall (68.27\% vs 67.37\%), F1 (74.05\% vs 73.45\%), OA (97.88\% vs 97.84\%), and IoU (58.79\% vs 58.04\%). The most notable improvements were observed in recall (+0.90\%) and IoU (+0.75\%), suggesting that the pixel-wise adversarial supervision particularly enhances the network's ability to identify and localize changes.
The experiment validates our idea that matrix-output discrimination is more appropriate for dense prediction tasks like CD, where the network needs to make decisions at each spatial location. The matrix-output design allows the discriminator to provide more granular feedback during adversarial training, helping the segmentation network better learn the spatial distribution of domain-invariant features at different locations.

\begin{table}[ht]\small
\caption{Results with Adversarial Domain Adaptation on LEVIR-CD $\rightarrow$ WHU-CD.}
\label{tab:ada2}
\centering
\begin{tabular}{cccccc}
\hline
\hline
Method & Precision & Recall & F1 & OA & IoU\\
\hline
$b1$ & 0.8073 & 0.6737 & 0.7345 & 0.9784 & 0.5804 \\
$b2$ & \textbf{0.8090} & \textbf{0.6827} & \textbf{0.7405} & \textbf{0.9788} & \textbf{0.5879} \\
\hline
\hline
\end{tabular}
\end{table}

Third, we conducted ablation studies to validate the effectiveness of our proposed alternating training strategy.
As analyzed in section \ref{secada}, balancing the weights between segmentation loss and adversarial loss poses a significant challenge in CD tasks.
Even when we introduced the multi-task adaptive weights by Kendall et al. \cite{kendall2018multi}, our experiments showed that the network still tends to overly focus on how to deceive the discriminator, leading to suboptimal convergence of the segmentation network and consequently compromising the CD performance.

\subsection{Experiment in Micro-Labeled Fine-Tuning}
\label{lMLFT}

This part of the experiment focuses on the our proposed sample selection strategy and the final training results.

First, we conducted comparative experiments on sample selection strategies with 16 labeled samples (approximately 0.3\% of the WHU-CD dataset).
Our DAM-Net selects samples based on their discriminability using the domain discriminator. We compared this approach with two baseline strategies:
\begin{itemize}
\item{$c1$: Random sample selection}
\item{$c2$: Selection based on detection performance}
\item{$c3$: Selection based on domain discriminator (DAM-Net)}
\end{itemize}
As shown in Table \ref{tab:sss}, our domain discriminator-based selection strategy ($c3$) achieves the best performance across all metrics (F1: 0.8309, IoU: 0.7108), significantly outperforming both detection performance-based selection ($c2$) and random selection ($c1$). Specifically, compared to detection performance-based selection ($c2$), our method improves precision by 4.88\%, recall by 1\%, F1 by 2.78\%, OA by 0.27\%, and IoU by 3.98\%. The performance gap becomes even larger when compared to random selection ($c1$).
These results reveal several important insights:
1) Samples with poor detection performance do not necessarily contribute more significantly to domain adaptation, although they still offer better guidance than randomly selected samples.
2) Our proposed strategy can effectively identify the most informative samples for both fine-tuning and domain adaptation tasks.
3) The superior performance of our selection strategy indirectly validates the effectiveness of the proposed ADA module, as such performance could only be achieved with a well-trained domain discriminator.

The experimental results convincingly demonstrate that our domain discriminator-based sample selection strategy can better identify critical samples for domain adaptation, leading to more effective model fine-tuning with limited labeled data.

\begin{table}[ht]\small
\caption{Results with Sample Selection Strategies on WHU-CD.}
\label{tab:sss}
\centering
\begin{tabular}{cccccc}
\hline
\hline
Method & Precision & Recall & F1 & OA & IoU\\
\hline
$c1$ & 0.8020 & 0.7389 & 0.7692 & 0.9807 & 0.6249 \\
$c2$ & 0.8263 & 0.7811 & 0.8031 & 0.9833 & 0.6710 \\
$c3$ & \textbf{0.8751} & \textbf{0.7911} & \textbf{0.8309} & \textbf{0.9860} & \textbf{0.7108} \\
\hline
\hline
\end{tabular}
\end{table}

\begin{figure*}[!ht]
\centering
\subfloat[]{
    \includegraphics[width= 0.15\textwidth]{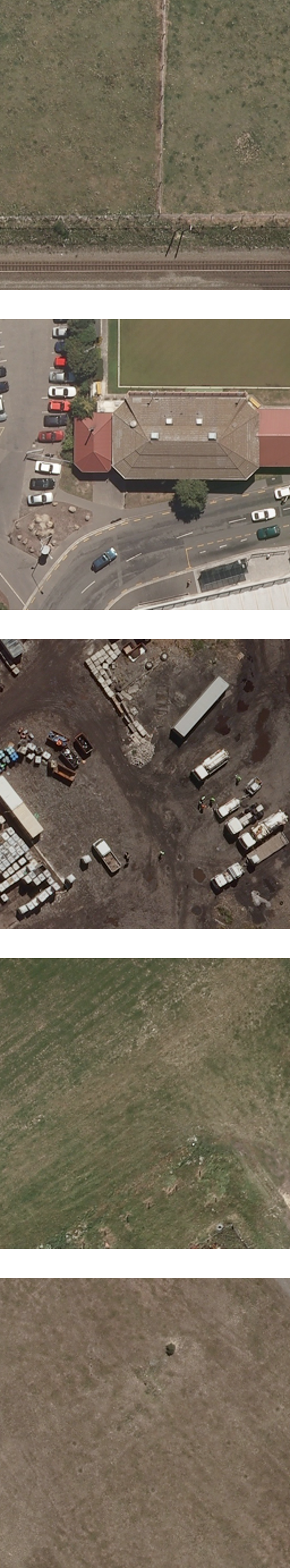}
}
\subfloat[]{
    \includegraphics[width= 0.15\textwidth]{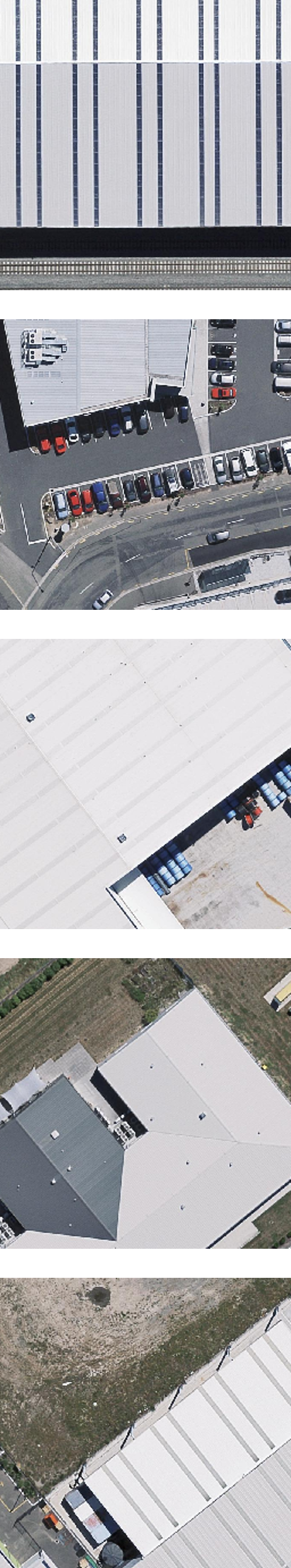}
}
\subfloat[]{
    \includegraphics[width= 0.15\textwidth]{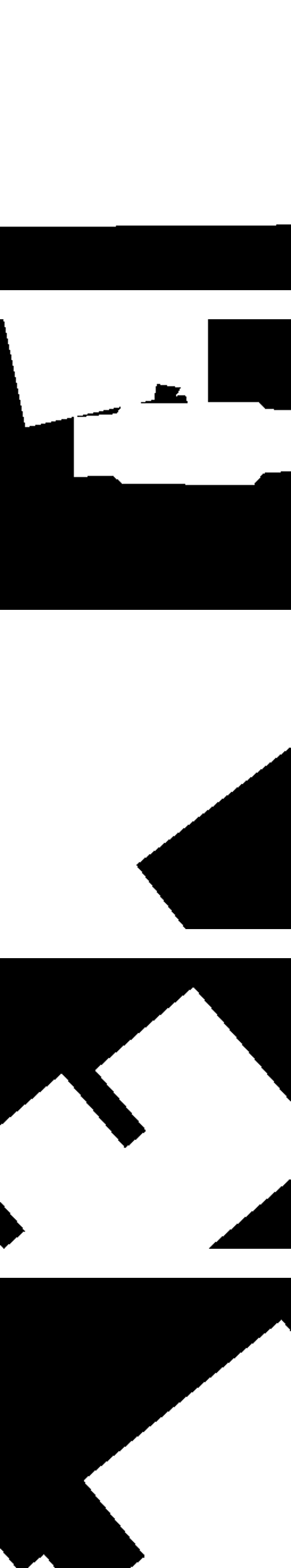}
}
\subfloat[]{
    \includegraphics[width= 0.15\textwidth]{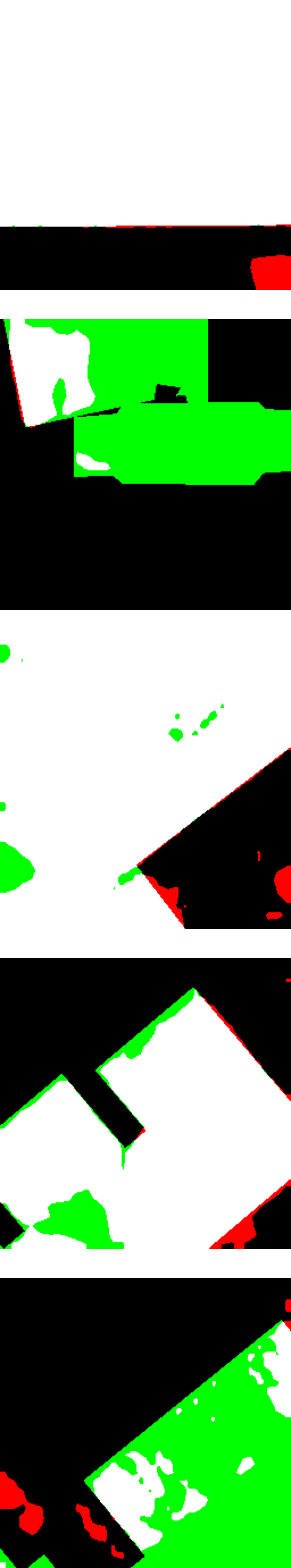}
}
\subfloat[]{
    \includegraphics[width= 0.15\textwidth]{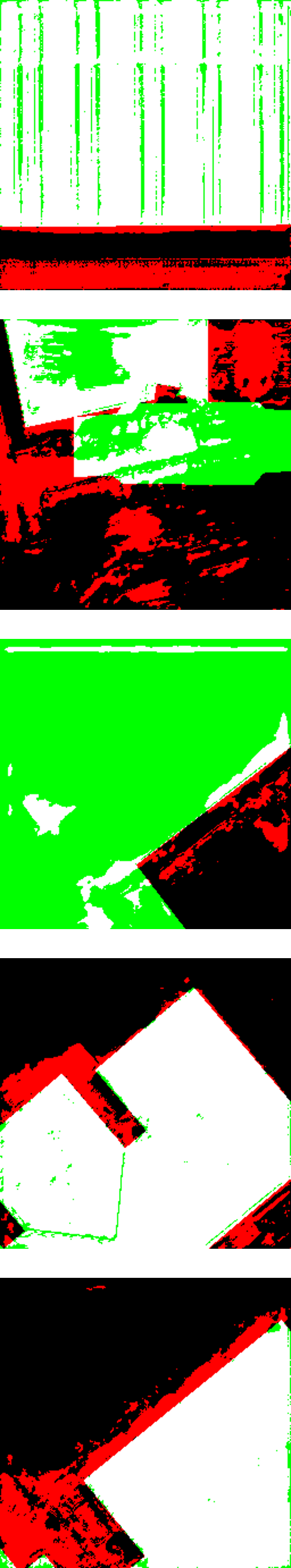}
}
\subfloat[]{
    \includegraphics[width= 0.15\textwidth]{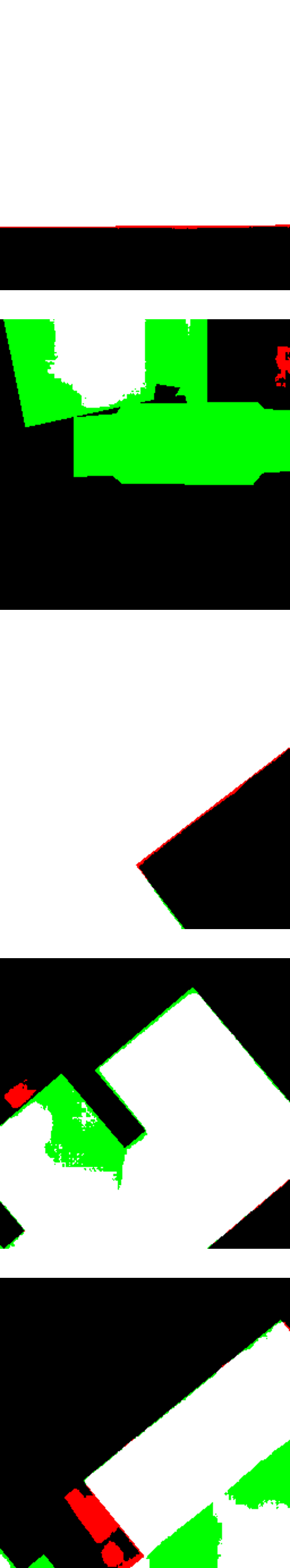}
}

\caption{(a)-(f) are results on WHU-CD Dataset (For semi-supervised methods, the entire LEVIR-CD dataset with full annotations alongside the WHU-CD dataset with 0.3\% of labeled samples are used as training dataset).
(a), (b) are the original images.
(c) is the ground truth.
The results of (d) SemiCD,
(e) C2F-SemiCD,
(f) our DAM-Net.
The false positives and false negatives are indicated in red and green, respectively.
Other colors represent true positives.}
\label{Semi}
\end{figure*}

Second, since DAM-Net incorporates MLFT that requires only a small number of labeled samples, we compared our method with state-of-the-art semi-supervised CD methods.
Specifically, we selected SemiCD proposed by Bandara \textit{et al.} \cite{bandara2022revisiting}, which proposed a simple yet effective way to leverage the information from unlabeled bi-temporal images, and C2F-SemiCD proposed by Han \textit{et al.} \cite{10445496}, which proposed a coarse-to-fine semi-supervised CD method based on consistency regularization.
Our DAM-Net only requires 16 labeled samples for fine-tuning, representing approximately 0.3\% of the WHU-CD dataset.
For fair comparison, we trained the semi-supervised methods using the entire LEVIR-CD dataset with full annotations alongside the WHU-CD dataset with varying proportions (0.3\%, 5\%, and 10\%) of labeled samples.
The results are presented in Table \ref{tab:final}.
Fig. \ref{Semi} illustrates representative detection results.
The false positives and false negatives are indicated by red and green, respectively. Other colors represent true positives.
The experimental results demonstrate several key findings:
1) With only 0.3\% labeled samples, DAM-Net achieves remarkable performance (F1: 0.8309, IoU: 0.7108), significantly outperforming both SemiCD (F1: 0.7644, IoU: 0.6186) and C2F-SemiCD (F1: 0.6921, IoU: 0.5292). 2) Most notably, DAM-Net with just 0.3\% labeled data achieves comparable or even better performance than semi-supervised methods using 10\% labeled samples.
3) The performance improvement of DAM-Net from 0.3\% to 10\% labeled data is relatively modest (F1: 0.8309 to 0.8407), indicating that our MLFT strategy is particularly effective with limited labeled samples.
These results convincingly demonstrate that our proposed DAM-Net, which synergistically combines domain adaptation with micro-labeled fine-tuning, can effectively compete with or surpass semi-supervised approaches while requiring significantly fewer labeled samples. This advantage makes DAM-Net particularly valuable in practical applications where labeled data is scarce or expensive to obtain.
\begin{table*}[ht]
\caption{Results with Semi-Supervised method on WHU-CD.}
\label{tab:final}
\centering
\begin{threeparttable}
\begin{tabular}{cccccccccc}
\hline
\hline
\multirow{2}*{Method} & \multicolumn{3}{c}{0.3\%} & \multicolumn{3}{c}{5\%} & \multicolumn{3}{c}{10\%} \\
\cline{2-10}
& F1 & IoU & OA & F1 & IoU & OA & F1 & IoU & OA \\
\hline
SemiCD & 0.7644 & 0.6186 & 0.9798 & 0.7977 & 0.6635 & 0.9815 & 0.8336 & 0.7147 & 0.9856 \\
C2F-SemiCD & 0.6921 & 0.5292 & 0.9736 & 0.7967 & 0.6621 & 0.9835 & 0.8154 & 0.6884 & 0.9849 \\
DAM-Net & \textbf{0.8309} & \textbf{0.7108} & \textbf{0.9860} & \textbf{0.8371} & \textbf{0.7198} & \textbf{0.9860} & \textbf{0.8407} & \textbf{0.8160} & \textbf{0.9880} \\
\hline
\hline
\end{tabular}
\begin{tablenotes}
\footnotesize
\item[*] For semi-supervised methods, the entire LEVIR-CD dataset with full annotations alongside the WHU-CD dataset with varying proportions (0.3\%, 5\%, and 10\%) of labeled samples are used as training dataset.
\end{tablenotes}
\end{threeparttable}
\end{table*}

\section{Discussion}
In this section, we discuss an interesting phenomenon observed in our experiments regarding the domain gap between LEVIR-CD and WHU-CD datasets. Our preliminary studies found the gap between these two widely-used CD benchmarks - while a network trained on LEVIR-CD achieved acceptable performance on WHU-CD with an F1 exceeding 50\%, the reverse scenario yielded substantially degraded results with an F1 below 30\%. This observation aligns with findings reported in several studies \cite{bandara2022revisiting}. Through systematic analysis, we identified two primary factors:
(1) the spatial resolution discrepancy between datasets, as evidenced by improved cross-dataset performance after resolution standardization, and (2) the presence of more complex scenarios in LEVIR-CD, particularly those involving tree occlusions, which are largely absent in WHU-CD. This complexity gap is reflected in the low recall values when adapting networks from WHU-CD to LEVIR-CD.
We hope these findings could provide valuable insights for researchers working on domain adaptation in CD tasks.

\section{Conclusion}
In this paper, we propose DAM-Net, a novel domain-adaptive method for CD that effectively addresses the domain shift challenge in remote sensing applications. Our work makes significant contributions to the field:
First, we introduce an innovative adversarial domain adaptation framework specifically designed for CD tasks. Unlike previous approaches that treat domain adaptation and CD separately, DAM-Net provides an end-to-end solution that holistically addresses both challenges. The effectiveness of this approach is demonstrated by substantial improvements over existing SOTA methods, achieving increases of 7.18\% in precision, 12.05\% in recall, and 10.29\% in F1 for the LEVIR-CD $\rightarrow$ WHU-CD adaptation task.
Second, we propose a Micro-Labeled Fine-Tuning (MLFT) strategy that significantly enhances network performance with minimal labeled data requirements. With just 16 labeled samples (about 0.3\% of the target dataset), MLFT further improves the network's performance, leading to additional gains of 6.61\% in precision, 10.84\% in recall, and 9.04\% in F1. Most notably, our approach with 0.3\% labeled data achieves comparable or superior performance to semi-supervised methods using 10\% labeled samples, demonstrating its exceptional efficiency in utilizing limited labeled data.
These results demonstrate that DAM-Net represents a significant advancement in domain adaption CD task, particularly in scenarios with limited labeled data availability. The synergistic combination of adversarial domain adaptation and micro-labeled fine-tuning provides a practical solution for real-world applications where labeled data is scarce or expensive to obtain. To facilitate further research in this direction, we will release our complete implementation, including source code and documents.

\section*{Acknowledgments}

Part of the numerical calculations in this paper have been done on the supercomputing system in the Supercomputing Center of Wuhan University.

\section*{Appendix}

\subsection{Multi-Temporal Transformer}

For the input multi-temporal features $x_1, x_2 \in {\mathbb R}^{h\times w\times c}$, they are first normalized through a normalization layer to obtain $x_1'$, $x_2'$, where $h$ and $w$ denote the height and width of the input feature map, respectively, and $c$ denotes the depth of the input feature maps.
The features describing the changes $f \in {\mathbb R}^{h\times w\times c}$ are obtained by $x_1'$ and $x_2'$.
Then $x_1'$, $x_2'$ and $f$ are passed through two linear layers ${\rm Linear}_Q$ and ${\rm Linear}_V$ to obtain the Query and Value matrices for the subsequent self-attention mechanism, the process can be formulated as
\begin{equation}
Q^*={\rm Linear}_Q\left(f\right)
\end{equation}
\begin{equation}
V_1={\rm Linear}_V\left(x_1'\right), V_2={\rm Linear}_{V}\left(x_2'\right)
\end{equation}
\begin{equation}
V={\rm Concat}(V_1,V_2)
\end{equation}
where ${\rm Linear}_Q$ and ${\rm Linear}_V$ refer to two linear layers with $c$ as input dimension and $c$ as output dimension, ${\rm Concat}()$ refers to concatenating matrices along the last dimension, and $V_1, V_2 \in {\mathbb R}^{h\times w\times c}$ represent the Value matrices corresponding to the multi-temporal inputs respectively, $Q^* \in {\mathbb R}^{h\times w\times c}$ represents the Query matrix without position embedding, $V \in {\mathbb R}^{h\times w\times 2c}$ represents the Value matrix.

Then, we add position embedding to the $Q^*, V_1, V_2$ matrices to obtain the $Q$ and $K$ matrices.
In this paper, we propose the spatio-temporal position embedding (STPE) which allows self-attentive mechanisms to perceive the temporal dimension of the input multi-temporal data better.
In transformers, global attention compares each token to all others, which is computationally expensive for large images \cite{dosovitskiy2020image}. Window attention \cite{liu2021swin} offers an alternative, comparing tokens only within local windows, thereby reducing computational complexity and making attention a function of window size rather than image size.
Bolya \textit{et al.} \cite{bolya2023window}suggested that there are bugs with the existing window position embedding, and they defined a separate window embedding and global embedding.
We use it as a spatial position embedding and and incorporate the temporal dimension, constructing spatio-temporal position embedding, which can be formulated as
\begin{equation}
P(i)=P_{window}+P_{global}+P_{temporal}(i)
\end{equation}
where $i$ represents the temporal dimension.

For the above $Q^*, V_1, V_2$ matrices (${\mathbb R}^{h\times w\times c}$), after dividing into windows, the dimension of the matrices will be rearanged, which can be expressed as $h\times w\times c = (hn\times wh)\times (wn\times ww)\times c\rightarrow hn\times wn\times wh\times ww\times c$,
where $h=hn*wh$, $w=wn*ww$, $hn$ and $wn$ represent the number of windows along the height and width, $wh$ and $ww$ represent the height and width of the window.
$P_{window}$, $P_{global}$ and $P_{temporal}^i$ are learnable parameter matrices,
the dimension of $P_{window}$ is $1\times 1\times wh\times ww\times c$, the dimension of $P_{global}$ is $hn\times wn\times 1\times 1\times c$, $P_{temporal}=\left\{P_{temporal}^f, P_{temporal}^1, P_{temporal}^2\right\}$ represents the set of temporal position embedding, where $P_{temporal}^f$, $P_{temporal}^1$ and $P_{temporal}^2$ represent the position embedding for different temporal inputs, and their dimensions are all $1\times 1\times 1\times 1\times c$.
Then, the spatio-temporal position embedding is added to $Q^*, V_1, V_2$ matrices to obtain the $Q$ (${\mathbb R}^{h\times w\times c}$) and $K$ (${\mathbb R}^{h\times w\times 2c}$) matrices, the process can be formulated as
\begin{equation}
Q=Q^*+P_{temporal}^f
\end{equation}
\begin{equation}
K_1=V_1+P_{temporal}^1, K_2=V_2+P_{temporal}^2
\end{equation}
\begin{equation}
K={\rm Concat}(K_1,K_2)
\end{equation}

Subsequently, the $K$, $Q$ and $V$ matrices are fed into the self-attention mechanism, and then the fusion results are obtained through the normalization and linear layers, which can be formulated as
\begin{equation}
F={\rm Linear}_A\left({\rm Norm}\left({\rm Attn}(Q, K, V)\right)\right)+{\rm Attn}(Q, K, V)
\end{equation}
where ${\rm Attn}()$ refers to the self-attention mechanism with $Q\in {\mathbb R}^{h\times w\times c}$, $K\in {\mathbb R}^{h\times w\times 2c}$, and $V\in {\mathbb R}^{h\times w\times 2c}$ as the inputs.
${\rm Norm}()$ refers to a normalization layer, ${\rm Linear}_A()$ refers to a linear layer with $c$ as the input and output dimension.




\bibliography{reference}

\begin{thebibliography}{10}
\providecommand{\url}[1]{#1}
\csname url@samestyle\endcsname
\providecommand{\newblock}{\relax}
\providecommand{\bibinfo}[2]{#2}
\providecommand{\BIBentrySTDinterwordspacing}{\spaceskip=0pt\relax}
\providecommand{\BIBentryALTinterwordstretchfactor}{4}
\providecommand{\BIBentryALTinterwordspacing}{\spaceskip=\fontdimen2\font plus
\BIBentryALTinterwordstretchfactor\fontdimen3\font minus \fontdimen4\font\relax}
\providecommand{\BIBforeignlanguage}[2]{{%
\expandafter\ifx\csname l@#1\endcsname\relax
\typeout{** WARNING: IEEEtran.bst: No hyphenation pattern has been}%
\typeout{** loaded for the language `#1'. Using the pattern for}%
\typeout{** the default language instead.}%
\else
\language=\csname l@#1\endcsname
\fi
#2}}
\providecommand{\BIBdecl}{\relax}
\BIBdecl

\bibitem{singh1989review}
A.~Singh, ``{Review Article Digital Change Detection Techniques using Remotely-Sensed Data},'' \emph{International Journal of Remote Sensing}, vol.~10, no.~6, pp. 989--1003, 1989.

\bibitem{bruzzone2012novel}
L.~Bruzzone and F.~Bovolo, ``A novel framework for the design of change-detection systems for very-high-resolution remote sensing images,'' \emph{Proceedings of the IEEE}, vol. 101, no.~3, pp. 609--630, 2012.

\bibitem{yang2021da2net}
R.~Yang, F.~Pu, Z.~Xu, C.~Ding, and X.~Xu, ``{DA2Net: Distraction-Attention-Driven Adversarial Network for Robust Remote Sensing Image Scene Classification},'' \emph{IEEE Geoscience and Remote Sensing Letters}, 2021.

\bibitem{li2020amn}
X.~Li, F.~Pu, R.~Yang, R.~Gui, and X.~Xu, ``{AMN: Attention Metric Network for One-Shot Remote Sensing Image Scene Classification},'' \emph{Remote Sensing}, vol.~12, no.~24, p. 4046, 2020.

\bibitem{lei2019landslide}
T.~Lei, Y.~Zhang, Z.~Lv, S.~Li, S.~Liu, and A.~K. Nandi, ``Landslide inventory mapping from bitemporal images using deep convolutional neural networks,'' \emph{IEEE Geoscience and Remote Sensing Letters}, vol.~16, no.~6, pp. 982--986, 2019.

\bibitem{peng2020optical}
X.~Peng, R.~Zhong, Z.~Li, and Q.~Li, ``Optical remote sensing image change detection based on attention mechanism and image difference,'' \emph{IEEE Transactions on Geoscience and Remote Sensing}, vol.~59, no.~9, pp. 7296--7307, 2020.

\bibitem{ji2018fully}
S.~Ji, S.~Wei, and M.~Lu, ``Fully convolutional networks for multisource building extraction from an open aerial and satellite imagery data set,'' \emph{IEEE Transactions on geoscience and remote sensing}, vol.~57, no.~1, pp. 574--586, 2018.

\bibitem{xian2010updating}
G.~Xian and C.~Homer, ``Updating the 2001 national land cover database impervious surface products to 2006 using landsat imagery change detection methods,'' \emph{Remote sensing of environment}, vol. 114, no.~8, pp. 1676--1686, 2010.

\bibitem{song2014remote}
C.~Song, B.~Huang, L.~Ke, and K.~S. Richards, ``Remote sensing of alpine lake water environment changes on the tibetan plateau and surroundings: A review,'' \emph{ISPRS Journal of Photogrammetry and Remote Sensing}, vol.~92, pp. 26--37, 2014.

\bibitem{wu2023fully}
C.~Wu, B.~Du, and L.~Zhang, ``Fully convolutional change detection framework with generative adversarial network for unsupervised, weakly supervised and regional supervised change detection,'' \emph{IEEE Transactions on Pattern Analysis and Machine Intelligence}, vol.~45, no.~8, pp. 9774--9788, 2023.

\bibitem{chen2019change}
H.~Chen, C.~Wu, B.~Du, L.~Zhang, and L.~Wang, ``Change detection in multisource vhr images via deep siamese convolutional multiple-layers recurrent neural network,'' \emph{IEEE Transactions on Geoscience and Remote Sensing}, vol.~58, no.~4, pp. 2848--2864, 2019.

\bibitem{li2022transunetcd}
Q.~Li, R.~Zhong, X.~Du, and Y.~Du, ``Transunetcd: A hybrid transformer network for change detection in optical remote-sensing images,'' \emph{IEEE Transactions on Geoscience and Remote Sensing}, vol.~60, pp. 1--19, 2022.

\bibitem{lecun1998gradient}
Y.~LeCun, L.~Bottou, Y.~Bengio, and P.~Haffner, ``Gradient-based learning applied to document recognition,'' \emph{Proceedings of the IEEE}, vol.~86, no.~11, pp. 2278--2324, 1998.

\bibitem{vaswani2017attention}
A.~Vaswani, ``Attention is all you need,'' \emph{Advances in Neural Information Processing Systems}, 2017.

\bibitem{gu2023mamba}
A.~Gu and T.~Dao, ``Mamba: Linear-time sequence modeling with selective state spaces,'' \emph{arXiv preprint arXiv:2312.00752}, 2023.

\bibitem{gu2022parameterization}
A.~Gu, K.~Goel, A.~Gupta, and C.~R{\'e}, ``On the parameterization and initialization of diagonal state space models,'' \emph{Advances in Neural Information Processing Systems}, vol.~35, pp. 35\,971--35\,983, 2022.

\bibitem{elman1990finding}
J.~L. Elman, ``Finding structure in time,'' \emph{Cognitive science}, vol.~14, no.~2, pp. 179--211, 1990.

\bibitem{hochreiter1997long}
S.~Hochreiter, ``Long short-term memory,'' \emph{Neural Computation MIT-Press}, 1997.

\bibitem{lecun2015deep}
Y.~LeCun, Y.~Bengio, and G.~Hinton, ``Deep learning,'' \emph{nature}, vol. 521, no. 7553, pp. 436--444, 2015.

\bibitem{hussain2013change}
M.~Hussain, D.~Chen, A.~Cheng, H.~Wei, and D.~Stanley, ``Change detection from remotely sensed images: From pixel-based to object-based approaches,'' \emph{ISPRS Journal of photogrammetry and remote sensing}, vol.~80, pp. 91--106, 2013.

\bibitem{lambin1994change}
E.~F. Lambin and A.~H. Strahlers, ``Change-vector analysis in multitemporal space: A tool to detect and categorize land-cover change processes using high temporal-resolution satellite data,'' \emph{Remote sensing of environment}, vol.~48, no.~2, pp. 231--244, 1994.

\bibitem{marchesi2009ica}
S.~Marchesi and L.~Bruzzone, ``Ica and kernel ica for change detection in multispectral remote sensing images,'' in \emph{2009 IEEE International Geoscience and Remote Sensing Symposium}, vol.~2.\hskip 1em plus 0.5em minus 0.4em\relax IEEE, 2009, pp. II--980.

\bibitem{shi2020change}
W.~Shi, M.~Zhang, R.~Zhang, S.~Chen, and Z.~Zhan, ``Change detection based on artificial intelligence: State-of-the-art and challenges,'' \emph{Remote Sensing}, vol.~12, no.~10, p. 1688, 2020.

\bibitem{hypersigma}
D.~Wang, M.~Hu, Y.~Jin, Y.~Miao, J.~Yang, Y.~Xu, X.~Qin, J.~Ma, L.~Sun, C.~Li, C.~Fu, H.~Chen, C.~Han, N.~Yokoya, J.~Zhang, M.~Xu, L.~Liu, L.~Zhang, C.~Wu, B.~Du, D.~Tao, and L.~Zhang, ``Hypersigma: Hyperspectral intelligence comprehension foundation model,'' \emph{arXiv preprint arXiv:2406.11519}, 2024.

\bibitem{zhu2017deep}
X.~X. Zhu, D.~Tuia, L.~Mou, G.-S. Xia, L.~Zhang, F.~Xu, and F.~Fraundorfer, ``Deep learning in remote sensing: A comprehensive review and list of resources,'' \emph{IEEE geoscience and remote sensing magazine}, vol.~5, no.~4, pp. 8--36, 2017.

\bibitem{guo2024skysense}
X.~Guo, J.~Lao, B.~Dang, Y.~Zhang, L.~Yu, L.~Ru, L.~Zhong, Z.~Huang, K.~Wu, D.~Hu \emph{et~al.}, ``Skysense: A multi-modal remote sensing foundation model towards universal interpretation for earth observation imagery,'' in \emph{Proceedings of the IEEE/CVF Conference on Computer Vision and Pattern Recognition}, 2024, pp. 27\,672--27\,683.

\bibitem{wang2018deep}
M.~Wang and W.~Deng, ``Deep visual domain adaptation: A survey,'' \emph{Neurocomputing}, vol. 312, pp. 135--153, 2018.

\bibitem{sun2016deep}
B.~Sun and K.~Saenko, ``Deep coral: Correlation alignment for deep domain adaptation,'' in \emph{Computer Vision--ECCV 2016 Workshops: Amsterdam, The Netherlands, October 8-10 and 15-16, 2016, Proceedings, Part III 14}.\hskip 1em plus 0.5em minus 0.4em\relax Springer, 2016, pp. 443--450.

\bibitem{tuia2016domain}
D.~Tuia, C.~Persello, and L.~Bruzzone, ``Domain adaptation for the classification of remote sensing data: An overview of recent advances,'' \emph{IEEE geoscience and remote sensing magazine}, vol.~4, no.~2, pp. 41--57, 2016.

\bibitem{goodfellow2020generative}
I.~Goodfellow, J.~Pouget-Abadie, M.~Mirza, B.~Xu, D.~Warde-Farley, S.~Ozair, A.~Courville, and Y.~Bengio, ``Generative adversarial networks,'' \emph{Communications of the ACM}, vol.~63, no.~11, pp. 139--144, 2020.

\bibitem{tsai2018learning}
Y.-H. Tsai, W.-C. Hung, S.~Schulter, K.~Sohn, M.-H. Yang, and M.~Chandraker, ``Learning to adapt structured output space for semantic segmentation,'' in \emph{Proceedings of the IEEE conference on computer vision and pattern recognition}, 2018, pp. 7472--7481.

\bibitem{roy2022uncertainty}
S.~Roy, M.~Trapp, A.~Pilzer, J.~Kannala, N.~Sebe, E.~Ricci, and A.~Solin, ``Uncertainty-guided source-free domain adaptation,'' in \emph{European conference on computer vision}.\hskip 1em plus 0.5em minus 0.4em\relax Springer, 2022, pp. 537--555.

\bibitem{chen2022rdpnet}
H.~Chen, F.~Pu, R.~Yang, R.~Tang, and X.~Xu, ``Rdp-net: Region detail preserving network for change detection,'' \emph{IEEE Transactions on Geoscience and Remote Sensing}, vol.~60, pp. 1--10, 2022.

\bibitem{chen2024srcnet}
H.~Chen, X.~Xu, and F.~Pu, ``Src-net: Bi-temporal spatial relationship concerned network for change detection,'' \emph{IEEE Journal of Selected Topics in Applied Earth Observations and Remote Sensing}, pp. 1--13, 2024.

\bibitem{hu2023a2dwqpe}
Y.~Hu, F.~Pu, C.~Kong, R.~Yang, H.~Chen, and X.~Xu, ``A2dwqpe: Adaptive and automated data-driven water quality parameter estimation,'' \emph{Journal of Hydrology}, vol. 626, p. 130363, 2023.

\bibitem{he2024visual}
Y.~He, X.~Xu, H.~Chen, J.~Li, and F.~Pu, ``Visual global-salient guided network for remote sensing image-text retrieval,'' \emph{IEEE Transactions on Geoscience and Remote Sensing}, 2024.

\bibitem{zhu2017change}
Z.~Zhu, ``Change detection using landsat time series: A review of frequencies, preprocessing, algorithms, and applications,'' \emph{ISPRS Journal of Photogrammetry and Remote Sensing}, vol. 130, pp. 370--384, 2017.

\bibitem{daudt2018urban}
R.~C. Daudt, B.~Le~Saux, A.~Boulch, and Y.~Gousseau, ``Urban change detection for multispectral earth observation using convolutional neural networks,'' in \emph{IGARSS 2018-2018 IEEE International Geoscience and Remote Sensing Symposium}.\hskip 1em plus 0.5em minus 0.4em\relax Ieee, 2018, pp. 2115--2118.

\bibitem{daudt2018fully}
R.~C. Daudt, B.~Le~Saux, and A.~Boulch, ``Fully convolutional siamese networks for change detection,'' in \emph{2018 25th IEEE international conference on image processing (ICIP)}.\hskip 1em plus 0.5em minus 0.4em\relax IEEE, 2018, pp. 4063--4067.

\bibitem{peng2019end}
D.~Peng, Y.~Zhang, and H.~Guan, ``End-to-end change detection for high resolution satellite images using improved unet++,'' \emph{Remote Sensing}, vol.~11, no.~11, p. 1382, 2019.

\bibitem{fang2021snunet}
S.~Fang, K.~Li, J.~Shao, and Z.~Li, ``Snunet-cd: A densely connected siamese network for change detection of vhr images,'' \emph{IEEE Geoscience and Remote Sensing Letters}, vol.~19, pp. 1--5, 2021.

\bibitem{zhou2018unet++}
Z.~Zhou, M.~M. Rahman~Siddiquee, N.~Tajbakhsh, and J.~Liang, ``Unet++: A nested u-net architecture for medical image segmentation,'' in \emph{Deep Learning in Medical Image Analysis and Multimodal Learning for Clinical Decision Support: 4th International Workshop, DLMIA 2018, and 8th International Workshop, ML-CDS 2018, Held in Conjunction with MICCAI 2018, Granada, Spain, September 20, 2018, Proceedings 4}.\hskip 1em plus 0.5em minus 0.4em\relax Springer, 2018, pp. 3--11.

\bibitem{papadomanolaki2021deep}
M.~Papadomanolaki, M.~Vakalopoulou, and K.~Karantzalos, ``A deep multitask learning framework coupling semantic segmentation and fully convolutional lstm networks for urban change detection,'' \emph{IEEE Transactions on Geoscience and Remote Sensing}, vol.~59, no.~9, pp. 7651--7668, 2021.

\bibitem{chen2021remote}
H.~Chen, Z.~Qi, and Z.~Shi, ``Remote sensing image change detection with transformers,'' \emph{IEEE Transactions on Geoscience and Remote Sensing}, vol.~60, pp. 1--14, 2021.

\bibitem{zhou2023mining}
F.~Zhou, C.~Xu, R.~Hang, R.~Zhang, and Q.~Liu, ``Mining joint intraimage and interimage context for remote sensing change detection,'' \emph{IEEE Transactions on Geoscience and Remote Sensing}, vol.~61, pp. 1--12, 2023.

\bibitem{chang2024remote}
H.~Chang, P.~Wang, W.~Diao, G.~Xu, and X.~Sun, ``Remote sensing change detection with bitemporal and differential feature interactive perception,'' \emph{IEEE Transactions on Image Processing}, 2024.

\bibitem{zhu2017unpaired}
J.-Y. Zhu, T.~Park, P.~Isola, and A.~A. Efros, ``Unpaired image-to-image translation using cycle-consistent adversarial networks,'' in \emph{Proceedings of the IEEE international conference on computer vision}, 2017, pp. 2223--2232.

\bibitem{vega2021unsupervised}
P.~J.~S. Vega, G.~A. O.~P. da~Costa, R.~Q. Feitosa, M.~X.~O. Adarme, C.~A. de~Almeida, C.~Heipke, and F.~Rottensteiner, ``An unsupervised domain adaptation approach for change detection and its application to deforestation mapping in tropical biomes,'' \emph{ISPRS Journal of Photogrammetry and Remote Sensing}, vol. 181, pp. 113--128, 2021.

\bibitem{saha2020unsupervised}
S.~Saha, Y.~T. Solano-Correa, F.~Bovolo, and L.~Bruzzone, ``Unsupervised deep transfer learning-based change detection for hr multispectral images,'' \emph{IEEE Geoscience and Remote Sensing Letters}, vol.~18, no.~5, pp. 856--860, 2020.

\bibitem{wittich2021appearance}
D.~Wittich and F.~Rottensteiner, ``Appearance based deep domain adaptation for the classification of aerial images,'' \emph{ISPRS Journal of Photogrammetry and Remote Sensing}, vol. 180, pp. 82--102, 2021.

\bibitem{chen2020dsdanet}
H.~Chen, C.~Wu, B.~Du, and L.~Zhang, ``Dsdanet: Deep siamese domain adaptation convolutional neural network for cross-domain change detection,'' \emph{arXiv preprint arXiv:2006.09225}, 2020.

\bibitem{song2019domain}
S.~Song, H.~Yu, Z.~Miao, Q.~Zhang, Y.~Lin, and S.~Wang, ``Domain adaptation for convolutional neural networks-based remote sensing scene classification,'' \emph{IEEE Geoscience and Remote Sensing Letters}, vol.~16, no.~8, pp. 1324--1328, 2019.

\bibitem{elshamli2017domain}
A.~Elshamli, G.~W. Taylor, A.~Berg, and S.~Areibi, ``Domain adaptation using representation learning for the classification of remote sensing images,'' \emph{IEEE Journal of Selected Topics in Applied Earth Observations and Remote Sensing}, vol.~10, no.~9, pp. 4198--4209, 2017.

\bibitem{ryali2023hiera}
C.~Ryali, Y.-T. Hu, D.~Bolya, C.~Wei, H.~Fan, P.-Y. Huang, V.~Aggarwal, A.~Chowdhury, O.~Poursaeed, J.~Hoffman \emph{et~al.}, ``Hiera: A hierarchical vision transformer without the bells-and-whistles,'' in \emph{International Conference on Machine Learning}.\hskip 1em plus 0.5em minus 0.4em\relax PMLR, 2023, pp. 29\,441--29\,454.

\bibitem{he2022masked}
K.~He, X.~Chen, S.~Xie, Y.~Li, P.~Doll{\'a}r, and R.~Girshick, ``Masked autoencoders are scalable vision learners,'' in \emph{Proceedings of the IEEE/CVF conference on computer vision and pattern recognition}, 2022, pp. 16\,000--16\,009.

\bibitem{woo2023convnext}
S.~Woo, S.~Debnath, R.~Hu, X.~Chen, Z.~Liu, I.~S. Kweon, and S.~Xie, ``Convnext v2: Co-designing and scaling convnets with masked autoencoders,'' in \emph{Proceedings of the IEEE/CVF Conference on Computer Vision and Pattern Recognition}, 2023, pp. 16\,133--16\,142.

\bibitem{sohn2020fixmatch}
K.~Sohn, D.~Berthelot, N.~Carlini, Z.~Zhang, H.~Zhang, C.~A. Raffel, E.~D. Cubuk, A.~Kurakin, and C.-L. Li, ``Fixmatch: Simplifying semi-supervised learning with consistency and confidence,'' \emph{Advances in neural information processing systems}, vol.~33, pp. 596--608, 2020.

\bibitem{yang2023revisiting}
L.~Yang, L.~Qi, L.~Feng, W.~Zhang, and Y.~Shi, ``Revisiting weak-to-strong consistency in semi-supervised semantic segmentation,'' in \emph{Proceedings of the IEEE/CVF Conference on Computer Vision and Pattern Recognition}, 2023, pp. 7236--7246.

\bibitem{chen2020spatial}
H.~Chen and Z.~Shi, ``A spatial-temporal attention-based method and a new dataset for remote sensing image change detection,'' \emph{Remote Sensing}, vol.~12, no.~10, p. 1662, 2020.

\bibitem{loshchilov2017decoupled}
I.~Loshchilov, ``Decoupled weight decay regularization,'' \emph{arXiv preprint arXiv:1711.05101}, 2017.

\bibitem{wang2024sfda}
J.~Wang and C.~Wu, ``Sfda-cd: A source-free unsupervised domain adaptation for vhr image change detection,'' \emph{Remote Sensing}, vol.~16, no.~7, p. 1274, 2024.

\bibitem{kendall2018multi}
A.~Kendall, Y.~Gal, and R.~Cipolla, ``Multi-task learning using uncertainty to weigh losses for scene geometry and semantics,'' in \emph{Proceedings of the IEEE conference on computer vision and pattern recognition}, 2018, pp. 7482--7491.

\bibitem{bandara2022revisiting}
W.~G.~C. Bandara and V.~M. Patel, ``Revisiting consistency regularization for semi-supervised change detection in remote sensing images,'' \emph{arXiv preprint arXiv:2204.08454}, 2022.

\bibitem{10445496}
C.~Han, C.~Wu, M.~Hu, J.~Li, and H.~Chen, ``C2f-semicd: A coarse-to-fine semi-supervised change detection method based on consistency regularization in high-resolution remote sensing images,'' \emph{IEEE Transactions on Geoscience and Remote Sensing}, vol.~62, pp. 1--21, 2024.

\bibitem{dosovitskiy2020image}
A.~Dosovitskiy, ``An image is worth 16x16 words: Transformers for image recognition at scale,'' \emph{arXiv preprint arXiv:2010.11929}, 2020.

\bibitem{liu2021swin}
Z.~Liu, Y.~Lin, Y.~Cao, H.~Hu, Y.~Wei, Z.~Zhang, S.~Lin, and B.~Guo, ``Swin transformer: Hierarchical vision transformer using shifted windows,'' in \emph{Proceedings of the IEEE/CVF international conference on computer vision}, 2021, pp. 10\,012--10\,022.

\bibitem{bolya2023window}
D.~Bolya, C.~Ryali, J.~Hoffman, and C.~Feichtenhofer, ``Window attention is bugged: How not to interpolate position embeddings,'' \emph{arXiv preprint arXiv:2311.05613}, 2023.

\end{thebibliography}
\bibliographystyle{IEEEtran}

\vfill

\end{document}